\documentclass[lettersize,journal]{IEEEtran}
\usepackage{amssymb}
\usepackage{placeins}
\usepackage{todonotes}
\usepackage{color}
\usepackage{hyperref}
\usepackage{mathtools}
\usepackage{makecell}
\usepackage{booktabs}
\usepackage{multirow} 
\usepackage{array}
\usepackage{subcaption}
\usepackage{textcomp}
\usepackage{stfloats}
\usepackage{dsfont}
\usepackage{url}
\usepackage{float}
\usepackage{verbatim}
\usepackage{graphicx}
\usepackage{amsmath,amsfonts,amsthm}
\usepackage{algorithmic}
\usepackage{algorithm}
\usepackage{graphicx}
\usepackage{cite}

\newtheorem{lemma}{Lemma}

\newtheorem{theorem}{Theorem}

\newtheorem{assumption}{Assumption}

\newtheorem{proposition}{Proposition}

\newtheorem{remark}{Remark}
\allowdisplaybreaks % Allow equations to break across pages
\begin{document}
\title{Convolutional Unscented Kalman Filter for Multi-Object Tracking with Outliers}

\author{Shiqi Liu, Wenhan Cao, Chang Liu, Tianyi Zhang, Shengbo Eben Li
\thanks{All correspondence should be sent to
Shengbo Eben Li.}
\thanks{Shiqi Liu is with the School of Vehicle and Mobility, Tsinghua University, Beijing, 100084, China. {Email:
lsq23@mails.tsinghua.edu.cn
}.}% <-this % stops a space
\thanks{Wenhan Cao is with the State Key Laboratory of Intelligent Green Vehicle and Mobility, Tsinghua University, Beijing, 100084, China. {Email:
cwh19@mails.tsinghua.edu.cn}.}
\thanks{Chang Liu is with the Department of Advanced Manufacturing and Robotics, Peking University, Beijing 100871, China. 
{Email: changliucoe@pku.edu.cn}.}
\thanks{Tianyi Zhang is with the School of Vehicle and Mobility, Tsinghua University, Beijing, 100084, China. 
{Email: 19241041@buaa.edu.cn}.}
\thanks{Shengbo Eben Li is with the School of Vehicle and Mobility and College of Artificial Intelligence, Tsinghua University, Beijing, 100084, China. {Email: lisb04@gmail.com}.}
}

% \markboth{Proceedings of the 2024 IEEE Conference on Decision and Control (CDC), December 2024}%
\markboth{Journal of \LaTeX\ Class Files,~Vol.~14, No.~8, August~2021}%
{Shell \MakeLowercase{\textit{et al.}}: A Sample Article Using IEEEtran.cls for IEEE Journals}


\maketitle

\begin{abstract}
Multi-object tracking (MOT) is an essential technique for navigation in autonomous driving. In tracking-by-detection systems, biases, false positives, and misses, which are referred to as outliers, are inevitable due to complex traffic scenarios.  Recent tracking methods are based on filtering algorithms that overlook these outliers, leading to reduced tracking accuracy or even loss of the object’s trajectory. To handle this challenge, we adopt a probabilistic perspective, regarding the generation of outliers as misspecification between the actual distribution of measurement data and the nominal measurement model used for filtering. We further demonstrate that, by designing a convolutional operation, we can mitigate this misspecification. Incorporating this operation into the widely used unscented Kalman filter (UKF) in commonly adopted tracking algorithms, we derive a variant of the UKF that is robust to outliers, called the convolutional UKF (ConvUKF). We show that ConvUKF maintains the Gaussian conjugate property, thus allowing for real-time tracking. We also prove that ConvUKF has a bounded tracking error in the presence of outliers, which implies robust stability. The experimental results on the KITTI and nuScenes datasets show improved accuracy compared to representative baseline algorithms for MOT tasks.

\end{abstract}


\section{Introduction}
\IEEEPARstart{M}{ulti-object} tracking (MOT) is an essential technology for autonomous driving \cite{luo2021multiple,liu2023efficient,li2023reinforcement}. This technique provides dynamic environment information for decision-making by
continuously monitoring the motion of surrounding objects. 
Currently, the majority of MOT algorithms adhere to the tracking-by-detection (TBD) paradigm \cite{weng2019baseline,guo20223d,li2023poly, wang2023camo, li2024fast}, which consists of three phases: object detection, data association, and filtering. The object detection phase uses a deep neural network to process input data from sensors, such as LiDAR and cameras, to identify objects and determine their motion states, represented as bounding boxes. Subsequently, the data association phase matches these detected bounding boxes with the predicted objects' trajectories, ensuring the continuous tracking of the same objects. Lastly, the filtering phase uses the matched detected bounding boxes as measurements to estimate the object's motion. 

\begin{figure}[htbp]
\centering
\begin{minipage}{0.24\textwidth}
\centering
\includegraphics[width=\linewidth]{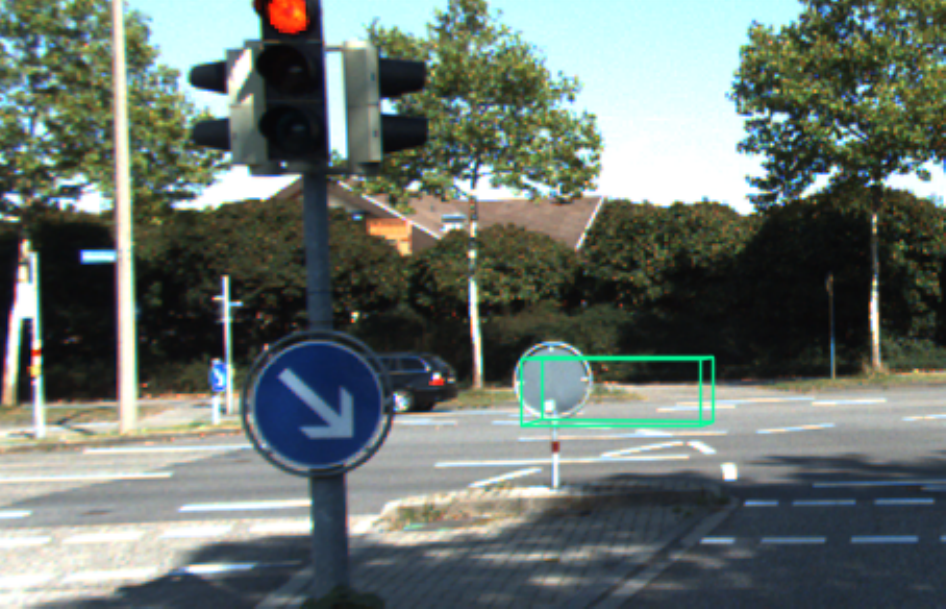}
\subcaption{}
\label{fig.outlier_1}
\end{minipage}
\begin{minipage}{0.24\textwidth}
\centering
\includegraphics[width=\linewidth]{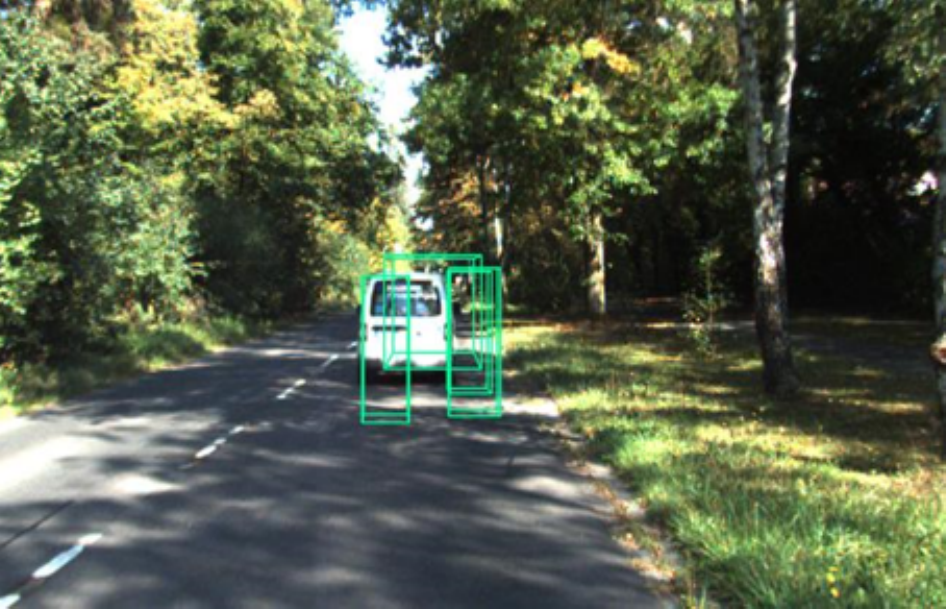}
\subcaption{}
\label{fig.outlier_2}
\end{minipage}
\qquad
\begin{minipage}{0.24\textwidth}
\centering
\includegraphics[width=\linewidth]{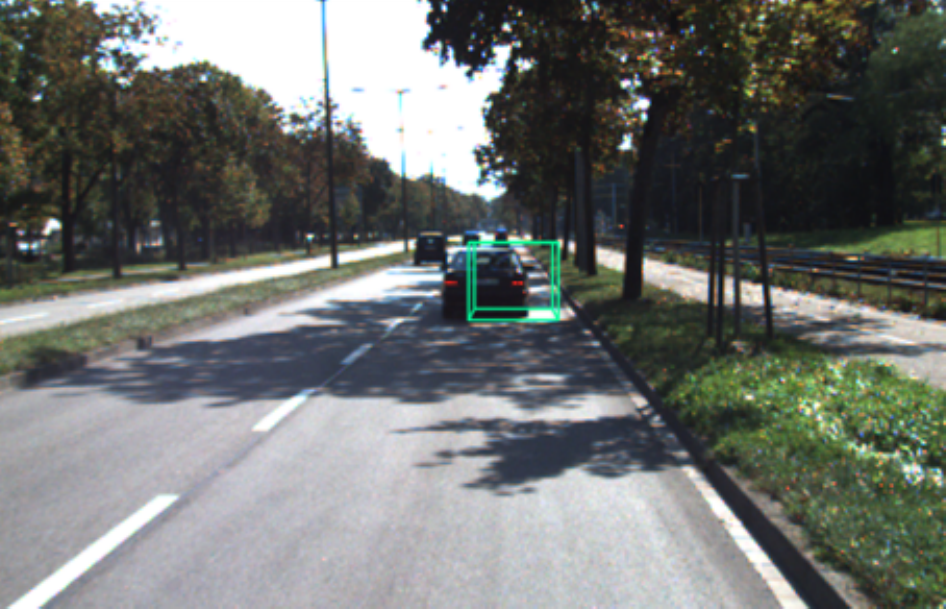}
\subcaption{}
\label{fig.outlier_3}
\end{minipage}
\begin{minipage}{0.24\textwidth}
\centering
\includegraphics[width=\linewidth]{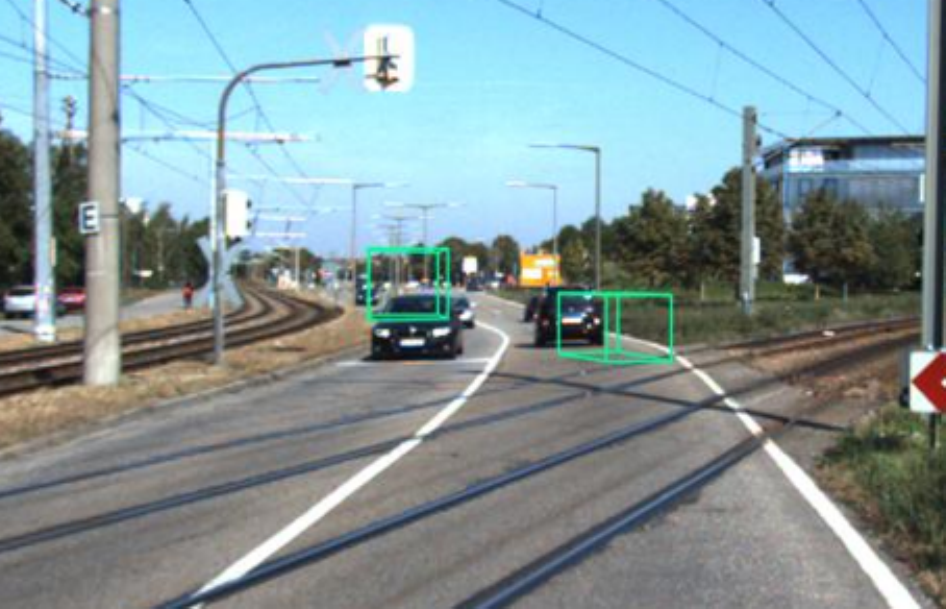}
\subcaption{}
\label{fig.outlier_4}
\end{minipage}
\caption{Examples of detection outliers from LiDAR dataset of KITTI \cite{geiger2013vision}. For better visualization, we project the detection results onto the image instead of using the original point cloud. Note that the bounding boxes are generated by running the PointRCNN detection algorithm \cite{shi2019pointrcnn}.  
(a) Detection misses: a black car is obscured by a signboard.
(b) False positives: redundant bounding boxes on the white car.
(c)(d) Detection bias: the bounding boxes  deviate from the desired car position.}
\label{fig.outliers}
\end{figure}
% \begin{figure}[!t]
% \centering
% \begin{subfigure}[b]{0.48\textwidth}
% \includegraphics[width=\textwidth]{figure/outlier_1.pdf}
% \captionsetup{font=footnotesize} 
% \caption{}
% \label{fig.outlier1}
% \end{subfigure}
% \hfill
% \begin{subfigure}[b]{0.48\textwidth}
% \includegraphics[width=\textwidth]{figure/outlier_2.pdf}
% \captionsetup{font=footnotesize}
% \caption{}
% \label{fig.outlier2}
% \end{subfigure}
% \caption{Visualization of outliers from the LiDAR point cloud of KITTI dataset \cite{geiger2013vision}, as detected by PointRCNN \cite{shi2019pointrcnn}. For better presentation, we visualize it on the image instead of the original point cloud. (a) Illustration of detection misses and false positives, exemplified by the black car obscured by a signboard.  (b) Highlight of significant bias.} 
% \label{fig.outliers}
% \end{figure}
The accuracy of MOT is significantly influenced by the matched bounding boxes from object detection and data association \cite{pal2021deep, rakai2022data}.
Practically, the complexity of cluttered environments and the limitation of detection and association algorithms often lead to a considerable number of biases, false positives, and false negatives (also called misses) \cite{dung2021robust, pang20213d}.
These occurrences lead to outliers, defined as the matched bounding boxes that differ significantly from others for the same objects \cite{cao2023generalized}.
An illustration of outliers in real-world scenarios is provided in Fig~\ref{fig.outliers}.
As input to the filtering phase, these outliers are challenging to model, inevitably causing misspecification of the nominal model and the actual measurement data, which can further deteriorate filtering performance.

Nonetheless, recent tracking methods are rooted in filtering algorithms that overlook the effects of outliers. For example, the most canonical and popular MOT algorithm, known as ABMOT3D \cite{weng2019baseline}, is based on the standard Kalman filter (KF). The standard KF naturally assumes that the underlying systems are linear and Gaussian, implying that there are no outliers in the measurement data. Following this line of research, recent works have extended this algorithm with more accurate nonlinear models for filtering to improve tracking performance, using extended KF \cite{omeragic2020tracking} and unscented KF \cite{liu2024ukf}. Unfortunately, these filtering techniques also assume Gaussian distributions for noise.
Under such conditions, outliers can cause deviations from Gaussian assumptions, leading to reduced accuracy and even the loss of tracking in MOT.
In an effort to tackle time-varying noises, the application of the adaptive cubature Kalman filter in MOT has been proposed \cite{ guo20223d}. However, the essence of this method is to adapt parameters to time-varying environments rather than to improve the algorithm's robustness under the contamination of outliers.

In this paper, we show that the matched bounding boxes in the TBD framework inevitably contain outliers, which would deteriorate the tracking performance when standard filtering algorithms are applied. 
In contrast to previous works that aim to avoid outliers by improving the robustness of detection \cite{li2022bevformer,qian20233d} and data association \cite{li2019robust, wang2023camo}, we focus on improving the filtering algorithm’s robustness
under the contamination of outliers.
We show that the misspecification between the nominal model and the measurement data caused by outliers can be quantitatively described using a threshold random variable to capture the uncertainty gap between them. We further demonstrate that this gap can be effectively incorporated through a specialized integral operation akin to convolution. 
Specifically, this operation aggregates the uncertainty gap throughout the filtering procedure, resulting in a filtering scheme that assigns lower weights to measurement outliers, thereby enhancing the robustness of the filtering.
Our contributions are summarized as follows:

\begin{figure*}[!htp]
\centering
\includegraphics[width=1.98\columnwidth]{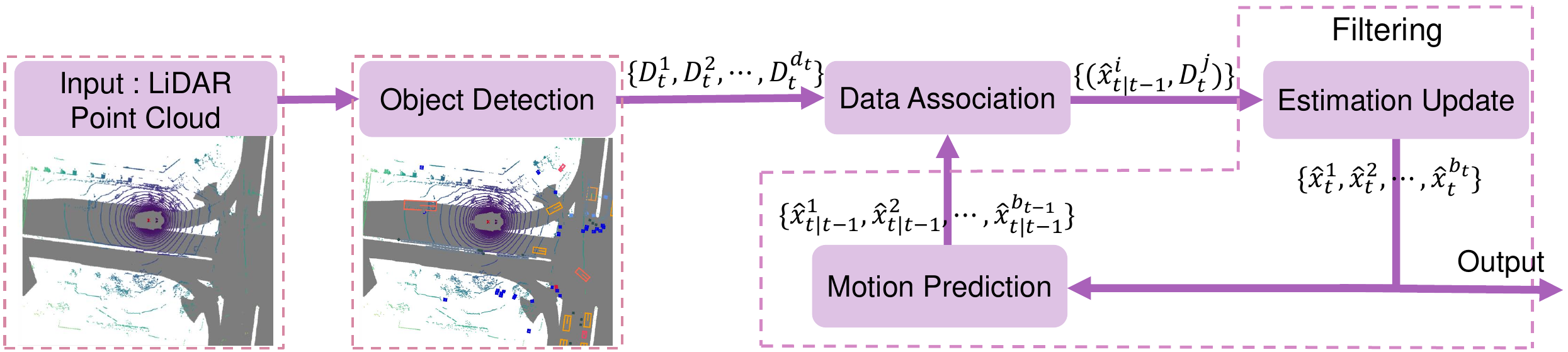}
\caption{TBD framework. Firstly, the object detection module detects bounding boxes $\{D_t^1,D_t^2,\dots,D_t^{d_t}\}$ from the raw LiDAR point cloud, providing initial information about the objects in the environment.
Next, the motion prediction in filtering module employs a motion model to predict the motions of detected objects, generating $\{\hat{x}_{t|t-1}^1,\hat{x}_{t|t-1}^2,\cdots,\hat{x}_{t|t-1}^{b_{t-1}}\}$.
Subsequently, the data association module matches these predictions with the current detections for the continuous tracking of the same objects $\{(\hat{x}^i_{t|t-1},D^j_t)\}$.
Finally, the estimation update in filtering module computes the motion estimation of objects $\{\hat{x}_{t}^{1},\hat{x}_{t}^{2},\cdots,\hat{x}_{t}^{b_{t}}\}$ at timestamp $t$.
As the next frame, object detection will process the incoming sensor data again.}
\label{fig.tracking}
\end{figure*}

\begin{itemize}

\item We demonstrate that the outliers can be modeled by introducing an uncertainty gap in the filtering model and mitigated by performing the so-called convolution operation in filtering algorithm. Applying this operation to the popular UKF algorithm results in a novel algorithm, which we refer to as the convolutional UKF (ConvUKF).

% \item 
% We analyze outliers using a threshold random variable to capture the uncertainty gap between the nominal
% model and the measurement data.
% We aggregate the uncertainty gap throughout the convolution operation in the UKF algorithm \cite{julier2004unscented}, resulting in lower weights to outliers.
% We name this innovative filter the Convolutional UKF (ConvUKF). 
% We propose a robust filtering algorithm called ConvUKF, specifically designed for filtering in MOT.
% By employing a convolution operation on the upper bound discrepancy between the real measurements, often contaminated by outliers, and the ideally modeled ones, 
% it reduces the weight of outliers in the calculation.
\item A stability proof for ConvUKF for nonlinear systems with measurement outliers is presented, showing that the estimation error is bounded in the mean square. Specifically, the upper bound of estimation error is proved to exhibit a linear positive correlation with both the initial state error and the covariance of noise and outliers.
\item The effectiveness of the proposed ConvUKF is validated using the real-world KITTI and nuScenes \cite{caesar2020nuscenes} datasets, which demonstrates an improvement in accuracy compared with baseline algorithms. Furthermore, our proposed method has the same computational complexity as UKF because it preserves the Gaussian conjugate structure, which enables real-time tracking.
% \cite{geiger2013vision} and nuScenes dataset \cite{caesar2020nuscenes}. It demonstrated an improvement of $18.42\%$ and $3.37\%$ in scaled Average Multi-Object Tracking Accuracy (sAMOTA), respectively, over the filtering baseline Huber Unscented Kalman Filter (HuberUKF)\cite{bing2018huber}. 
\end{itemize}

The structure of the remaining sections is structured as follows: 
Section \ref{sec.framework} presents the preliminaries of TBD framework and problem formulation of filtering. 
In Section \ref{sec.ConvUKF}, the ConvUKF algorithm and its stability analysis are proposed. 
% Section \ref{sec.discussion} delves into the discussion on existing robust filtering methods for long-tail noise or outliers. 
In Section \ref{sec.simulations}, we provide an evaluation on the KITTI and nuScenes datasets, along with comparisons with existing filtering methods. 
Finally, conclusions and limitation discussion are drawn in Section \ref{sec.conclusion}.

\textbf{Notation:}
The notation $M \geq N$ indicates that the matrix $M-N$ is positive semi-definite, while $M > N$ indicates that the matrix $M-N$ is positive definite. For vectors, $\| x \|$ represents the $l_2$-norm of the vector $x$. 
For matrices, $\|A\|$ means the Frobenius norm of the matrix $A$. 
% For $M\succeq0$, $\Vert x \Vert_M$ is short for $\sqrt{x^{\top}Mx}$. $\Vert M \Vert_{\mathcal{F}}$ is the Frobenius norm of matrix $M$ while $|M|$ signifies the determinant of the matrix $M$. 
Besides, $I_{m \times n}$ and $0_{m\times n}$ are the identity matrix and zero matrix  with dimension $m\times n$.
% $\|A\|$ means the Frobenius norm of the matrix $A$.

% \section{Problem Formulation}

\section{Preliminaries and Problem Formulation}\label{sec.framework}
The objective of MOT is to track surrounding objects using raw sensor data. The most applied TBD framework, illustrated in Fig~\ref{fig.tracking}, comprises three modules: object detection, data association, and filtering \cite{weng2019baseline,guo20223d,li2023poly,li2024fast}. 
In this section, we introduce the preliminaries of the TBD framework and present the problem formulation of filtering under this framework.
\subsection{Preliminaries}\label{subsec.Preliminaries}
\textbf{Object detection.} In MOT, object detection is aimed at identifying and locating the vehicle's surroundings. The process begins with raw data input from onboard sensors like cameras and LiDAR, which capture the scene around the vehicle. The raw data is then fed into a neural network that extracts features and identifies objects  \cite{pang20213d}.
The output of this module is a set of detected bounding boxes represented as $D_t=\{D_t^1,D_t^2,\dots,D_t^{d_t}\}$, where $d_t$ denotes the number of detections. Each detection, $D_t^i$ is described as a tuple $(p_x, p_y, p_z, \phi, l, w, h)$ for $i\in\{1,2,\dots,d_t\}$, which includes the position of geometric center $(p_x, p_y, p_z)$, size of detected bounding boxes $(l, w, h)$, and yaw angle $(\phi)$.

\textbf{Data association.} The data association module aligns the predicted trajectories $\{\hat{x}_{t|t-1}^{1},\hat{x}_{t|t-1}^{2},\cdots,\hat{x}_{t|t-1}^{b_{t-1}}\}$ provided by motion prediction part in filtering module, with the bounding boxes $D_t$, sourced from the detection module to get the appropriate match $\{(\hat{x}^i_{t|t-1},D^j_t)\}$, $i \in [1, b_{t-1}], \, j \in [1, d_t]$, as illustrated by Fig~\ref{fig.tracking}. Here $b_{t-1}$ is the number of predicted trajectories in the filtering module.
To accomplish the best match, this module encompasses the similarity $a_{ij}$ between each pair of predicted trajectory $\hat{x}_{t|t-1}^i$ and detection $D^j_t$:
% The similarity is calculated based on the 3D intersection over union (IoU) \cite{weng2019baseline},
% \Winstons{negative}
% which computes the negative distance between the centers of each trajectory $\hat{x}_{t|t-1}^i$ and detection $D^j_t$
% $\forall i \in [1, b_{t-1}], \, j \in [1, d_t]$,
\begin{equation}\nonumber
\begin{aligned}
% A = [a_{ij}],\;
a_{ij}=\frac{\|\hat{x}_{t|t-1}^i\bigcap D^j_t\|}{\|\hat{x}_{t|t-1}^i\bigcup D^j_t\|},\forall i \in [1, b_{t-1}], \, j \in [1, d_t],
\end{aligned}
\end{equation}
where $\|\hat{x}_{t|t-1}^i\bigcap D^j_t\|$ and $\|\hat{x}_{t|t-1}^i\bigcup D^j_t\|$ represent the intersection volume and the sum volume between the detected objects and the predicted objects, respectively.
Using this kind of similarity, the graph-matching problem be formulated and resolved in polynomial time using the Hungarian algorithm \cite{kuhn1955hungarian}.

\textbf{Filtering.}
The filtering module employs filtering algorithms to estimate objects' motion.
Given the necessity of computational efficiency for real-time tracking, the majority of methods employ the Kalman filter family \cite{weng2019baseline,li2018kalman, liu2024ukf, guo20223d},
which consists of two parts: motion prediction and estimation update.
The motion prediction part utilizes the estimated state $\{\hat{x}_{t-1}^{1},\hat{x}_{t-1}^{2},\cdots,\hat{x}_{t-1}^{b_{t-1}}\}$ from the previous frame ($t-1$) to forecast the movements of objects, resulting in predictions denoted as  $\{\hat{x}_{t|t-1}^{1},\hat{x}_{t|t-1}^{2},\cdots,\hat{x}_{t|t-1}^{b_{t-1}}\}$. 
These predictions will be fed into the data association module for further processing.
On the other hand, the estimation update part receives matched pairings $\{(\hat{x}^i_{t|t-1}, D^j_t)\}$ from the data association module.
With this input, it calculates the final motion estimations of objects $\{\hat{x}_{t}^{1},\hat{x}_{t}^{2},\cdots,\hat{x}_{t}^{b_{t}}\}$ through the filtering algorithms.

\subsection{Problem Formulation of Filtering}
In MOT, the objective of filtering is to mitigate noise and yield an accurate estimation of the object’s motion state $x = [p_x, p_y, p_z, \phi, l, w, h, v_h, v_v, \Dot{v}_h, \Dot{\phi}]^\top$, including position $(p_x, p_y, p_z)$, size $(l, w, h)$, horizontal velocity and vertical velocity $ (v_h, v_v)$, horizontal acceleration $\Dot{v}_h$, yaw angle $\phi$, yaw rate $\Dot{\phi}$. 
The horizontal velocity $v_h$ and the horizontal acceleration $\dot{v}_h$ are defined in the direction of the object's orientation, while the vertical velocity $v_v$ is along the z-axis.
The system is constructed as a state-space model:
\begin{subequations} \label{eq.ssm}
\begin{align}
x_{t+1}&=f\left(x_{t}\right)+\xi_{t}, \label{eq.ssm_transiton}\\
y_{t}&=h\left(x_{t}\right)+\zeta_{t},\label{eq.ssm_obs}
\end{align}
\end{subequations}
where $f(\cdot)$ and $h(\cdot)$ represent the state transition and measurement models, respectively. The term $x_t\in \mathcal{X}\subseteq R^n$ is defined as the object's motion state while $y_t\in \mathcal{Y}\subseteq R^m$ denotes the specific matched bounding boxes $D_t$. 
Furthermore, $\xi_t\sim \mathcal{N}(0, Q_t)$ represents the Gaussian transition noise, characterized by a covariance matrix $Q_t$. The measurement noise $\zeta_t$, under ideal circumstances, is assumed to follow a Gaussian distribution with $\zeta_t\sim \mathcal{N}(0, R_t)$ with the covariance $R_t$.
% \begin{remark}
% In complex real-world scenarios of MOT, identifying the appropriate noise parameters $Q_t$ and $R_t$ poses a significant challenge.
% In recent MOT methods, we divide the way to identify the parameter into two categories: empirical value \cite{weng2019baseline,li2023poly,li2024fast} and adaptive parameter adjustment \cite{guo20223d, hoffmann2020real}.
% The former employs an empirical value such as identity matrix $I$ as the covariance.
% The latter focuses on how to dynamically adjust the covariance in reality applications.
% \end{remark}

The transition model $f(\cdot)$ is to predict the motion of objects given the state $x_t$. For motion tracking, there are typically choices, namely the constant velocity (CV) model, the constant turn rate and velocity (CTRV) model, and the constant turn rate and acceleration (CTRA) model \cite{4632283}.
Among them, the CTRA model offers a most precise depiction of motion in real-world object-tracking scenarios by incorporating both turning and acceleration factors \cite{4632283},
whose formulation is shown as
\begin{equation}\nonumber
\begin{aligned}
f(x_{t})&=x_{t}\\
&+
[\Delta p_{x,t},\;
\Delta p_{y,t},\;
v_{v,t}\Delta t,\;
\Dot{\phi}_t\Delta t,\;
0_{1 \times 3},\;
\Dot{v}_{h,t}\Delta t,\;
0_{1 \times 3}]^\top ,\\
\Delta p_{x,t} &= \frac{1}{\Dot{\phi}_t^2}[(v_{h,t}+\Dot{v}_{h,t}\Delta t)\Dot{\phi}_t\sin{(\phi_t+\Dot{\phi}_t\Delta t)}\\
&-{v}_{h,t}\Dot{\phi}_t\sin{\phi_t}
+\Dot{v}_{h,t}\cos{(\phi_t+\Dot{\phi}_t\Delta t)}-\Dot{v}_{h,t}\cos{\phi_t}],\\
\Delta p_{y,t} &= \frac{1}{\Dot{\phi}_t^2}[(-v_{h,t}-\Dot{v}_{h,t}\Delta t)\Dot{\phi}_t\cos{(\phi_t+\Dot{\phi}_t\Delta t)}\\
&+{v}_{h,t}\Dot{\phi}_t\cos{\phi_t}
+\Dot{v}_{h,t}\sin{(\phi_t+\Dot{\phi}_t\Delta t)}-\Dot{v}_{h,t}\sin{\phi_t}], 
\end{aligned}
\end{equation}
where $\Delta t$ is the sample time.
As we introduced in Section \ref{subsec.Preliminaries}, measurements are derived from the matched bounding boxes, represented as $(p_x, p_y, p_z, \phi, l, w, h)$, thus the related measurement model can be formalized as
\begin{equation}\nonumber
\begin{aligned}
h(x_t)
&=\begin{bmatrix}
% I_{7 \times 7} &0 \\
% 0 & 0_{4\times4}
I_{7 \times 7}, 0_{4\times4}
\end{bmatrix}x_t.
\end{aligned}
\end{equation}

% To derive the updated tracked states from the matched pairings of predicted states and corresponding detected bounding boxes post-data association, many existing tracking methodologies conventionally employ the Kalman filter (KF) \cite{weng2019baseline} for models with linear transitions, or the Unscented Kalman Filter (UKF) \cite{nie20233d} for those with nonlinear transitions. Nonetheless, in real-world tracking scenarios, outliers can significantly impair the performance of these filtering algorithms, potentially causing estimation errors or even algorithm divergence, leading to tracking failures. To mitigate this issue, we introduce the ConvUKF, highlighting its role in improving tracking accuracy in the presence of outliers in the following section.

% Notably, the prediction state $\hat{x}_{t|t-1}$, crucial for data association, is calculated as \eqref{eq.x_predict}. And the output of MOT, denoted by the motion of objects $\hat{x}_t$, is derived using \eqref{eq.x_update}.
% The details of the ConvUKF will be introduced in the subsequent section.

\begin{remark}

The state space model (SSM) \eqref{eq.ssm} can also be represented as the  hidden Markov model (HMM):
\begin{equation} \label{eq.hmm}
\begin{aligned}
% {x}_0 &\sim p_0(x_0),\nonumber
% \\
{x}_t \sim p(x_t|x_{t-1}), \;
{y}_t \sim p(y_t|x_t),
\end{aligned}
\end{equation}
where $p(x_t|x_{t-1})$ represents the transition probability, and $p(y_t|x_t)$ is the likelihood probability.
Under the ideal conditions, we have $p(x_t|x_{t-1})=\mathcal{N}(x_t;f(x_{t-1}),Q_t)$ and $p(y_t|x_{t})=\mathcal{N}(y_t;h(x_{t}),R_t)$.
\end{remark}

\section{Convolutional Unscented Kalman Filter}\label{sec.ConvUKF}
The UKF, as one of the Kalman family filters presented by \cite{julier2004unscented}, stands out as a widely adopted method for its efficiency and high precision in estimating the state of nonlinear motion in MOT \cite{liu2024ukf}.
However, the UKF assumes an ideal HMM as described in \eqref{eq.hmm}, and the presence of outliers in measurements can significantly challenge its estimation performance.
In this section, we analyze the HMM with outliers and present the ConvUKF algorithm under this analysis. Finally, we demonstrate the stability of ConvUKF in nonlinear systems with outliers.
% ConvUKF involves measuring uncertainty in the model into the filtering framework.

\subsection{Algorithm Design} \label{algorithm.ConvUKF}
\begin{figure}
\centering
\includegraphics[width=0.99\linewidth]{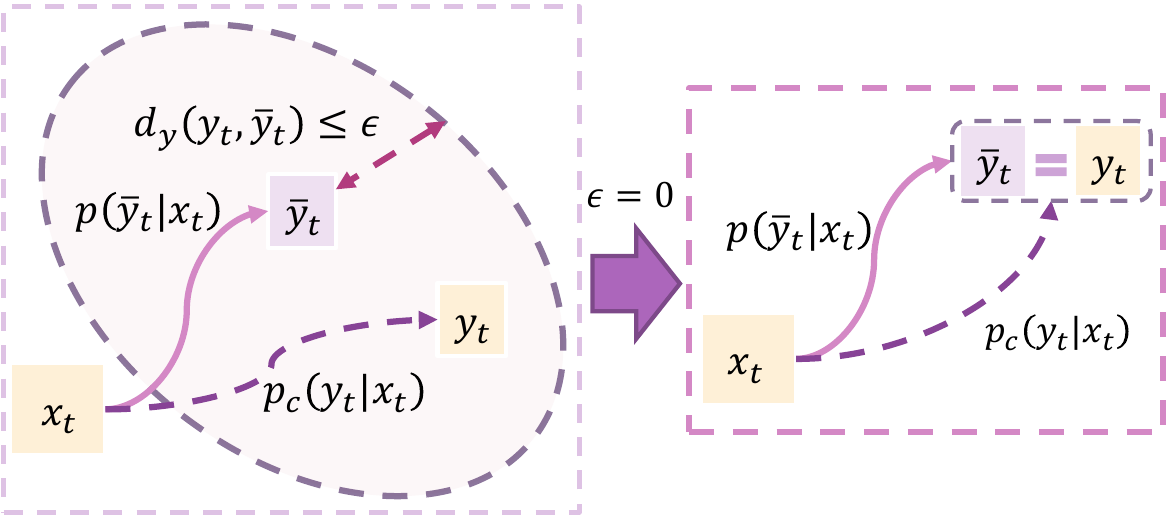}
\caption{Illustration of the HMM with outliers and without outliers. The nominal likelihood probability in HMM projects the state in the previous time, denoted as ${x}_{t-1}$, to a modeled virtual measurement $\bar{{y}}_t$. However, due to stochastic nature of outliers, there is an uncertainty gap between ${y}_t$ and $\bar{{y}}_t$. We assume their gap distance $d_y({y}_t, \bar{{y}}_t)$ can be modeled by a stochastic inequality ${\epsilon}$, i.e., $d_y({y}_t, \bar{{y}}_t) \leq {\epsilon}$. 
Especially when ${\epsilon}=0$, meaning no outliers, the system is reduced to the ideal scene $\bar{{y}}_t={y}_t$.}
\label{fig.rectified_HMM}
\end{figure}

In MOT, attaining precise model $p(y_t|x_t)$ in \eqref{eq.hmm} is often unfeasible due to the inevitable biases, false
positives, and false negatives \cite{dung2021robust, pang20213d}. 
Consequently, it is necessary to differentiate the actual measurement variable ${y}_t$, often contaminated by outliers, and the virtual measurement variable $\bar{{y}}_t$, which is shown in Fig~\ref{fig.rectified_HMM}.
The former is an accurate yet unattainable description of the measurement, while the latter is an artificial construct generated by nominal models.

Following research \cite{cao2024convolutional}, we construct a stochastic inequality to model the uncertainty gap between ${y}_t$ and $\bar{{y}}_t$
by a threshold random variable ${\epsilon}$, represented as
\begin{equation}\label{eq.stochastic equality}
d_y({y}_t, \bar{{y}}_t) \leq {\epsilon}
\end{equation}
Here, $d_y:\mathcal{Y}\times \mathcal{Y} \rightarrow R$ denotes the distance function. Note that the cumulative distribution function of the threshold random variable is assumed to be known and denoted as $F_{\epsilon}$.  Combined with the original HMM \eqref{eq.hmm}, this constitutes the HMM with outliers:
\begin{equation} \label{eq.rectified hmm}
\begin{aligned}
% &{x}_0 \sim p_0(x_0),\nonumber
% \\
{{x}}_t \sim p({x}_t|x_{t-1}),\;
\bar{{y}}_t \sim p(\bar{y}_t|x_t),\;
d_y({y}_t, \bar{{y}}_t) \leq {\epsilon}, \end{aligned}
\end{equation}
where $p(\bar{y}_t|x_t) = \mathcal{N}(\bar{y}_t;h(x_t),R_t)$ is the likelihood probability for the virtual measurement variable $\bar{y}_t$ under the ideal condition, called the nominal likelihood. 
Note that when ${\epsilon}=0$,  \eqref{eq.rectified hmm} can be reduced to the ideal HMM without outliers, as illustrated by Fig~\ref{fig.rectified_HMM}.
Due to the fact that the physical world and the modeling of a system are mutually exclusive at any given moment, it is reasonable to assume that $y_t$ and $\bar{y_t}$ are conditionally independent given $x_t$, and ${\epsilon}$ is also independent of both ${y}_t$ and $\bar{{y}}_t$\cite{cao2024convolutional}.

Under the formulation of HMM with outliers in \eqref{eq.rectified hmm}, we can calculate the so-called convolutional likelihood probability \cite{cao2024convolutional} by additionally conditioning on stochastic inequality \eqref{eq.stochastic equality}:
\begin{equation} \label{eq.convolution operation}
\begin{aligned}
p(y_t|x_{t}, d(y_t, \bar{y}_t) \leq \epsilon) \propto\int_{\bar{y}_t}\left(1-F_{\epsilon}(d_y(y_t,\bar{y}_t))\right)p(\bar{y}_t|x_{t})\mathrm{d}\bar{y}_t.
\end{aligned}    
\end{equation}
Here, $p_c(y_t|x_t) := p(y_t|x_{t}, d(y_t, \bar{y}_t) \leq \epsilon)$ is the convolutional likelihood. The convolutional likelihood owes its name as it is calculated by an integral operation akin to convolution, which integrates the uncertainty gap $\epsilon$ caused by outliers. Specifically, $1-F_{\epsilon}(d_y(y_t,\bar{y}_t))$ serves as the kernel function, which is
a weighting coefficient of the nominal likelihood $p(\bar{y}_t|x_t)$ based on the distance $d_y(y_t,\bar{y}_t)$. The convolutional likelihood is proved to be a more informative estimate of the actual likelihood function than the nominal one \cite{cao2024convolutional} when the system is contaminated by outliers. 
% Moreover, one can directly replace the nominal likelihood function in the update step of Kalman filter family to improve the robustness of the filter:
% \Winstons{Bayesian filter}
% \begin{equation}\label{eq.conv BF update}
% p(x_t|y_{1:t}) = \frac{p(x_t|y_{1:t-1})p_c(y_t|x_t)}{\int p(x_t|y_{1:t-1})p_c(y_t|x_t) \,\mathrm{d}x_{t}}.    
% \end{equation}
% We name \eqref{eq.conv BF update} as the convolutional update.

A practical challenge with this scheme is whether this newly defined likelihood has an analytical form, given that the convolution operation typically lacks a closed-form solution \cite{pei2000closed}. Fortunately, for the common quadratic-form distance, if the threshold random variable follows an exponential distribution, the nominal model with  Gaussian noise possesses an analytical form of the convolutional likelihood function. This is demonstrated in the subsequent lemma.
\begin{lemma}
[\cite{cao2024convolutional}]\label{lemma.Convolutional Bayesian Filtering}

Consider the following nominal system model
$p(\bar{y}_t|x_t) = \mathcal{N}(\bar{y}_t;h(x_t),R_t)$.
If $d_y({y}, \bar{{y}}) = \|{y} - \bar{{y}} \|^2$ and ${\epsilon} \sim \mathrm{Exp}(\gamma)$ with $\gamma>0$ as the parameter of the exponential distribution, then we have $p_c(y_t|x_{t}) = \mathcal{N}(y_t; g(x_t), R_t + \frac{1}{2\gamma} \cdot I_{m \times m})$.
\end{lemma}
As shown in this lemma, if the nomial likelihood function is Gaussian, its convolutional counterpart is still Gaussian. This property allows us to preserve the original Gaussian conjugate structure if we apply the convolutional update to the well-known KF filter family to which UKF belongs. 
By reshaping the original likelihood of UKF to a convolutional likelihood, we have the following ConvUKF algorithm:
\begin{itemize}
\item  
\textbf{Initialization.} Given: $\hat{x}_0, \hat{P}_0, Q, R, \gamma, a$, repeat steps $1\mathrm{-}3$ for $t=0, 1,2,\dots$.
\item 
% \textbf{Step 1: Select Julier sigma points \cite{julier2004unscented}.}   
\textbf{Step 1: Select sigma points.} 
Here are the Julier sigma points:
$\mathcal{X}_{0,t}=\hat{x}_t; \mathcal{X}_{i,t}=\hat{x}_t+(a\sqrt{n\hat{P}_t})_i; \mathcal{X}_{i,t}=\hat{x}_t-(a\sqrt{n\hat{P}_t})_i, i=1,2,\dots,n$, where $a$ is a proportion parameter and $(a\sqrt{n\hat{P}_t})_i$
is the vector of the ith column of the matrix square root.
\item \textbf{Step 2: Prediction.}
\begin{equation}
\label{eq.prediction}
\begin{aligned}
\hat{x}_{t+1|t}&=\sum_{i=0}^{2n} \omega_i f(\mathcal{X}_{i,t}), i=0,1,2,\dots,2n,  \\
\breve{\mathcal{X}}_{i,t+1|t}&=f(\mathcal{X}_{i,t})-\hat{x}_{t+1|t},\\
\hat{P}_{t+1|t} &=
\sum_{i=0}^{2n} \omega_i \breve{\mathcal{X}}_{i,t+1|t}
\breve{\mathcal{X}}_{i,t+1|t}^\top 
+ {Q}_t ,
\end{aligned}  
\end{equation}
where $\omega_0=1-\frac{1}{a^2},\omega_i=\frac{1}{2na^2}, i=1,2,\dots,2n$ are
the scalar weights with $\sum_{i=0}^{2n} \omega_i=1$. It is noticeable that $\hat{x}_{t+1|t}$ is the prediction state, and $\hat{P}_{t+1|t}$ is the predicted covariance at time step $t$. Both $\hat{x}_{t+1|t}$ and $\hat{P}_{t+1|t}$ will be corrected in the update step.
\item \textbf{Step 3: Update.}
\begin{subequations}
\begin{align}
\hat{y}_{t+1}&=\sum_{i=0}^{2n} \omega_i h(\mathcal{X}_{i,t}), , i=0,1,2,\dots,2n,\nonumber\\
\breve{\mathcal{Y}}_{i,t+1}&=h(\mathcal{X}_{i,t})-\hat{y}_{t+1},\nonumber\\
\hat{P}_{yy,t+1} &=
\sum_{i=0}^{2n} \omega_i \breve{\mathcal{Y}}_{i,t+1}
\breve{\mathcal{Y}}_{i,t+1}^\top+R_{t+1}+\frac{1}{2\gamma} \cdot I_{m \times m} ,\label{eq.P_yy_update}\\
\hat{P}_{xy,t+1} &=
\sum_{i=0}^{2n} \omega_i \breve{\mathcal{X}}_{i,t+1|t}\breve{\mathcal{Y}}_{i,t+1}^\top,\label{eq.P_xy_update}\\
K_{t+1} &= \hat{P}_{xy,t+1}\hat{P}_{yy,t+1}^{-1},\label{eq.K_update}\\
\hat{x}_{t+1}&=\hat{x}_{t+1|t}+K_{t+1}(y_{t+1}-\hat{y}_{t+1}),\label{eq.x_update}\\
\hat{P}_{t+1} &=\hat{P}_{t+1|t}-K_t\hat{P}_{xy,t+1}^\top \label{eq.P_update}.
\end{align}  
\end{subequations}

In \eqref{eq.P_yy_update} and \eqref{eq.P_xy_update} $\hat{P}_{yy,t+1}$ and $\hat{P}_{xy,t+1}$  represent the measurement covariance and the cross-covariance at time step $t+1$. 
The term $\frac{1}{2\gamma} \cdot I_{m \times m}$ in \eqref{eq.P_yy_update} is derived from the convolutional probability in Lemma \ref{lemma.Convolutional Bayesian Filtering}. Meanwhile, $\hat{x}_{t+1}$ and $\hat{P}_{t+1}$ denote the estimated state and its associated covariance, respectively.
% \item \textbf{Step 4: Repeat steps 1–3 for the next frame.}
\end{itemize}

In MOT, the parameter $\gamma$ in \eqref{eq.P_yy_update} is associated with the extent of outlier contamination in measurements, which is challenging to determine.

Inspired by the adaptive filtering algorithms in \cite{guo20223d}, we adopt a similar adaptive update rule for $\gamma$:

\begin{equation}
\label{eq.gamma_adaptive}
\begin{aligned}
\gamma^{(t+1)}&=(1-\tau)\gamma^{(t)}\\
&+
\tau\frac{\gamma^{(t)}}{1+\exp{\{-2\gamma^{(t)}[\exp{(-\gamma^{(t)})}-\frac{\tilde{y_t}}{m}]}\}},\\
\end{aligned}  
\end{equation}

where $\tau$, set to the empirical value of $0.05$ according to  \cite{guo20223d}, is the temperature parameter regulating the update rate, and $\tilde{y_t}=y_{t+1}-\hat{y}_{t+1}$ denotes the measurement error. The underlying concept is straightforward: as the measurement error increases, indicating a larger mismatch between the distribution of measurement data and its corresponding model, $\gamma$ should be reduced to accommodate this greater uncertainty gap. Conversely, if the error diminishes, the opposite adjustment will be made.

% \begin{remark}
% This adaptive update rule adopts the empirical work from \cite{guo20223d} to avoid complex parameter adjustments. The main contribution of this research is the integration of convolution operations into the UKF to handle outliers. The effectiveness of this approach is demonstrated through the subsequent ablation experiment.
% \end{remark}

\subsection{Stability Analysis}
To employ the proposed algorithm in MOT, it is necessary to ensure filtering stability in nonlinear systems with outliers. Mathematically, filtering stability requires that the estimation error, $\tilde{x}_{t+1} = x_{t+1} - \hat{x}_{t+1}$, is bounded in the mean square, i.e., $\mathbb{E}[\|\tilde{x}_{t+1}\|^2] < +\infty$ \cite{xiong2006performance}.
One challenge for stability analysis is that ConvUKF uses UT to approximate the nonlinear system with a linear system, which inevitably introduces approximation error. A viable solution to compensate for this approximation error is to introduce auxiliary variables in the linear approximation \cite{xiong2006performance,li2012stochastic}, leading to the following assumption:
\begin{assumption}
\label{asp.linear approximation}
(Rectified Linearization\cite{xiong2006performance, li2012stochastic})
For the nonlinear system with outliers in \eqref{eq.ssm}, the prediction error $\tilde{x}_{t+1|t}=x_{t+1}-\hat{x}_{t+1|t}$ and measurement error $\tilde{y}_{t+1}=y_{t+1}-\hat{y}_{t+1}$ can be linearly updated with the auxiliary variables $\alpha_t = \mathrm{diag}\{\alpha_{1,t},\dots,\alpha_{n,t}\}$ and 
$\beta_t = \mathrm{diag}\{\beta_{1,t},\dots,\beta_{n,t}\}$.
\begin{subequations}
\label{eq.model error}
\begin{align}
\tilde{x}_{t+1|t}&=\alpha_tF_t\tilde{x}_{t|t-1}-\alpha_tF_tK_t\tilde{y}_{t|t-1} + \xi_t,\label{eq.pred_x_error}\\
\tilde{y}_{t+1}&=\beta_{t+1}H_{t+1}\tilde{x}_{t+1|t}+ \zeta_{t+1},\label{eq.y_error}
\end{align}  
\end{subequations}
where $F_t=(\frac{\partial f(x)}{\partial x}|_{x=\hat{x}_t})$ and $H_{t+1}=(\frac{\partial h(x)}{\partial x}|_{x=\hat{x}_{t+1|t}})$ are Jacobian matrices.
\end{assumption}

Assumption \ref{asp.linear approximation} essentially assumes the neglect of higher-order terms when using UT can be rectified by the auxiliary variables $\alpha_t$ and $\beta_t$.
Based on this assumption, the predicted covariance $\hat{P}_{t+1|t}$ and the measurement covariance $\hat{P}_{yy,t+1}$ can be computed in the following proposition.

\begin{proposition}\label{prop.covariance_update}(Rectified Covariance Update\cite{li2012stochastic})
Under Assumption \ref{asp.linear approximation}, the predicted covariance $\hat{P}_{t+1|t}$ and the measurement covariance $\hat{P}_{yy,t+1}$ are derived as: 
\begin{subequations}
\begin{align}
\hat{P}_{t+1|t}&=[\alpha_tF_t(I-K_t\beta_tH_t)]\hat{P}_{t|t-1} \nonumber \\
&\times[\alpha_tF_t(I-K_t\beta_tH_t)]^\top+{Q}_t+\Delta P_{t+1|t},
\label{eq.hat_P}\\
\hat{P}_{yy,t+1}&=(\beta_{t+1}H_{t+1})\hat{P}_{t+1|t}(\beta_{t+1}H_{t+1})^\top \nonumber\\
&+R_{t+1}+\frac{1}{2\gamma} \cdot I_{m \times m}+\Delta P_{yy,t+1},
\label{eq.hat_Pyy}
\end{align}  
\end{subequations}
where $\Delta P_{t+1}$ and $\Delta P_{yy,t+1}$ denote the error covariance, capturing the discrepancies between the actual nonlinear dynamics and their linear approximations, defined as \eqref{eq.Delat_P}.
\begin{subequations}\label{eq.Delat_P}
\begin{align}
\Delta P_{t+1|t}&=\mathbb{E}\{[\alpha_tF_t(I-K_t\beta_tH_t)\tilde{x}_{t+1|t}]\nonumber\\
&\times
[\alpha_tF_t(I-K_t\beta_tH_t)\tilde{x}_{t+1|t}]^\top\}\nonumber\\
&-[\alpha_tF_t(I-K_t\beta_tH_t)]\hat{P}_{t|t-1}
[\alpha_tF_t(I-K_t\beta_tH_t)]\nonumber\\
&-P_{t+1|t} +\hat{P}_{t+1|t},
\label{eq.Delta_P_t}\\
\Delta P_{yy,t+1}&=\mathbb{E}[(\beta_{t+1}H_{t+1})\tilde{x}_{t+1|t}\tilde{x}_{t+1|t}^\top(\beta_{t+1}H_{t+1})^\top]\nonumber\\
&-(\beta_{t+1}H_{t+1})\hat{P}_{t+1|t}(\beta_{t+1}H_{t+1})^\top\nonumber\\
&-P_{yy,t+1} +\hat{P}_{yy,t+1}
\label{eq.Delta_P_yy}
\end{align}  
 \end{subequations}
\end{proposition}

The proof of Proposition \ref{prop.covariance_update} can be found in Appendix \ref{appendix.Proof of Proposition 1}.
This proposition implies that by adding the covariance error term, we can derive the calculation formula for the estimated covariance.
The covariance error term can be seen as the rectification covariance of noise.
Considering the presence of outliers, we define the nominal measurement covariance $R_t$ and the discrepancy between the true covariance and the nominal one induced by outliers as $\Delta R_t$.
Subsequently, we will perform a stability analysis under the assumption that certain variables within the system adhere to boundedness, as outlined below.
\begin{assumption} \label{asp.boundeness}
We assume that the following 3 parts of the hypothesis are satisfied for all $t>0$.
\begin{itemize}
\item \textbf{Bounded dynamics}:
The dynamics of the system are bounded by real constants ($f_{u}, h_{u}, \beta_{u}, \alpha_{u}$): 
\begin{equation}
\nonumber
\begin{aligned}
F_tF_t^T\leq f_{u}^2I,\; H_tH_t^T\leq h_{u}^2I,\;
\alpha_t\leq\alpha_{u}I,\; \beta_t \leq \beta_{u}I.
\end{aligned}  
\end{equation}
\item \textbf{Bounded noise}:
There are real constants ($q_{l}, q_{u}, {r}_{u}, \Delta {r}_{u}$) that bound the noise covariance:
\begin{equation}
\nonumber
\begin{aligned}
q_{l}I\leq Q_t\leq q_{u}I,\;
R_t\leq{r}_{u}I,\;
0\leq \Delta R_t\leq\Delta{r}_{u} I.
\end{aligned}  
\end{equation}
\item
\textbf{Bounded rectified covariance}.
For the error covariance, this assumption establishes bounds through real constants $\Delta {p}_{l},\Delta {p}_{u}, \Delta {p}_{yy,u}, {p}_{l}, {p}_{u}$:
\begin{equation}
\nonumber
\begin{aligned}
\Delta {p}_{l}I\leq\Delta P_{t+1|t} &\leq \Delta {p}_{u}I,\; 
\Delta P_{yy,t} \leq \Delta {p}_{yy,u}I,\\
{p}_{l}I&\leq\hat{P}_{t}\leq{p}_{u}I.
\end{aligned}  
\end{equation}
\end{itemize}
\end{assumption}

Assumption \ref{asp.boundeness} is a prerequisite for filtering stability \cite{reif1999stochastic,xiong2006performance,li2012stochastic}, assuming stable dynamics and finite noise conditions.
Based on Assumption
\ref{asp.boundeness}, we can derive
that the bound of the predicted covariance $\hat{P}_{t+1|t}$ and the norm of Kalman gain $\|K_{t+1}\|$.
\begin{table*}[!t]
\centering
\caption{\centering{\textsc{Performance Comparison on KITTI Validation Set.}}}
\label{tab:performance_comparison}
\begin{tabular}{@{}llcccccccc@{}}
\toprule
\textbf{Dataset} & \textbf{Method} & \textbf{sAMOTA$\% \uparrow$} & \textbf{AMOTA$\%\uparrow$} & \textbf{AMOTP$\%\uparrow$} & \textbf{MT$\%\uparrow$} & \textbf{ML$\%\downarrow$} & \textbf{IFN$\downarrow$} & \textbf{FRAG$\downarrow$} & \textbf{Time (ms)$\downarrow$} \\ 
\midrule
\multirow{5}{*}{KITTI} 
& UKF & 76.30& 31.47& 51.83& 47.57& 5.41& 898& 321& 0.4225 \\
& HuberUKF & 84.42& 38.12& 61.47& 69.19& 3.24& {725}& 50& 0.9496 \\
& AUKF& 84.63& 38.02& 74.80& 70.81& \textbf{2.16}& 843& 270& 0.6510\\
& IEKF & 83.30& 37.30& 58.31& 68.11& 4.86& 828& 69& \textbf{0.1890}\\
& REKF & 89.51& 42.15& 65.23& 71.89& 3.78& 792& 47& 0.3225\\
& ICKF & 88.34& 41.32& 45.50& 70.81& 3.24& \textbf{695}& 67& 0.7186 \\
& ConvUKF & \textbf{93.32}& \textbf{45.46}& \textbf{78.09}& \textbf{75.68} & 3.78& 799& \textbf{17}& 0.5726\\
\bottomrule
\multicolumn{10}{l}{The best performances are marked with \textbf{bold} font. The symbol $\uparrow$ indicates better performance for larger values.}
\end{tabular}
\end{table*}

\begin{table*}[!t]
\centering
\caption{\centering{\textsc{Performance Comparison on nuScenes Validation Set.}}}
\label{tab:performance_comparison_nuscenes}
\begin{tabular}{@{}llcccccccc@{}}
\toprule
\textbf{Dataset} & \textbf{Method} & \textbf{sAMOTA$\% \uparrow$} & \textbf{AMOTA$\%\uparrow$} & \textbf{AMOTP$\%\uparrow$} & \textbf{MT$\%\uparrow$} & \textbf{ML$\%\downarrow$} & \textbf{IDS$\downarrow$} & \textbf{FRAG$\downarrow$} & \textbf{Time (ms)$\downarrow$} \\ 
\midrule
\multirow{5}{*}{nuScenes}
& UKF & 62.66 & 21.93 & 32.07 & 37.90 & 28.29& 3973 & 4740 & 0.3102 \\
& HuberUKF & 66.13 & 23.69 & 39.05 & 39.67 & 30.09 & 799 & 1499 & 0.6024 \\
& AUKF & 67.15& 24.19& \textbf{53.86}& 41.49& \textbf{27.76}& 1959& 2520& 0.3138\\
& IEKF & 56.11& 18.02& 25.01& 33.46& 30.90& \textbf{662}& 1817& \textbf{0.1946}\\
& REKF & 64.27& 22.54& 33.03& 39.60& 29.19 & 721& 1440& 0.2906\\
& ICKF & 66.12& 23.50& 16.19& 38.34& 30.46 & 899& 1959& 0.7238\\
& ConvUKF & \textbf{69.12}& \textbf{25.05}& 49.58& \textbf{42.86}& 29.34& 703&  \textbf{1171}& 0.5788\\
\bottomrule
\end{tabular}
\end{table*}

\begin{proposition} \label{prop.P_tK_t_range}
The prediction covariance $\hat{P}_{t+1|t}$ and the norm of the Kalman gain matrix $\|K_{t+1}\|$ are  bounded as $\hat{p}_{l}I\leq \hat{P}_{t+1|t}\leq\hat{p}_{u}I$  and $\|K_{t+1}\|\leq K_u$  with $\hat{p}_{l}={p}_{l},
\hat{p}_{u}={p}_{u}\bar{a}^2f_{u}^2+q_{u}+\Delta{p}_{u},
K_u=\sqrt{n}({p}_{u}\alpha_u^{2}f_{u}^2+q_{u}+\Delta{p}_{u}).$
% \begin{equation}
% \label{eq.P_tK_t_range}
% \begin{aligned}
% \hat{p}_{l}&={p}_{l},\\
% \hat{p}_{u}&={p}_{u}\bar{a}^2f_{u}^2+q_{u}+\Delta{p}_{u},\\
% K_u&=\sqrt{n}{p}_{u}.
% \end{aligned}  
% \end{equation}
\end{proposition}
Proposition \ref{prop.P_tK_t_range} provides the bound of $\hat{P}_{t+1|t}$ and $\|K_{t+1}\|$ with proof of this proposition detailed in Appendix \ref{Appendix.Proof of Proposition 2}.
Then we can formulate and prove the following stability theorem for the ConvUKF:
\begin{theorem} \label{the.bound}
For nonlinear stochastic systems with outliers \eqref{eq.ssm}, the estimation error of ConvUKF is bounded in the mean square if Assumption \ref{asp.linear approximation} and \ref{asp.boundeness} are satisfied.
Specifically, the bound has a linear positive correlation with the initial state error and the upper bound of measurement noise and outliers' covariance.
\begin{equation}
\label{eq.x_bound_thm}
\begin{aligned}
\mathbb{E}\{\|\tilde{x}_{t+1}\|^2\}\leq C_1\mathbb{E}\{\|\tilde{x}_{0}\|^2\}
+C_2({r}_{u}+\Delta{r}_{u})+C_3,
\end{aligned}  
\end{equation}
where $C_1, C_2, C_3$ are real constants uncorrelated with $\tilde{x}_{0}$, ${r}_{u}$, $\Delta{r}_{u}$, $\{y_k\}_{k=1:t+1}$ and $\{\hat{x}_k\}_{k=1:t+1}$.
\end{theorem}
The proof of Theorem \ref{the.bound} and detailed definition 
 of $C_1, C_2, C_3$ can be found in Appendix \ref{appendix.Proof of Theorem 2}. 
The basic proof sketch for Theorem \ref{the.bound} is to construct a quadratic function of the error as a Lyapunov function to demonstrate stability. 
This technique is widely employed in previous works \cite{xiong2006performance,li2012stochastic}.
Notably, we consider the systems with measurement outliers, a factor overlooked in the aforementioned works.

\begin{remark}
From Theorem 1, we can deduce that the bound of estimation error is linear positively correlated with the initial state error and the upper bound of measurement noise and outliers' covariance. Specifically, the impact of various system parameters on the error bound of state estimation can be inferred through the definition of constants $C_1$, $C_2$, and $C_3$ in \eqref{eq.constant_bound}.
For example, the estimation error of a system with lower $f_u$ will be less affected by outliers. 
\end{remark}
 
% Specifically, when the upper bound of outlier caused covariance error $\gamma\xleftarrow \infty$ , the theorem reduces to the stability analysis of ConvUKF for common dynamic systems \cite{cao2024convolutional}. 
% \begin{remark}
% The basic proof sketch for Theorem \ref{the.bound} is to construct a quadratic function of the error as a Lyapunov function to demonstrate stability. 
% This technique is widely employed in previous works \cite{xiong2006performance,li2012stochastic,huang2020novel}.
% Notably, we consider the systems with measurement outliers, a factor overlooked in the aforementioned works.
% \end{remark}

\section{Experiments}\label{sec.simulations}

\begin{figure*}[!htp]
\centering
\includegraphics[width=2.0\columnwidth]{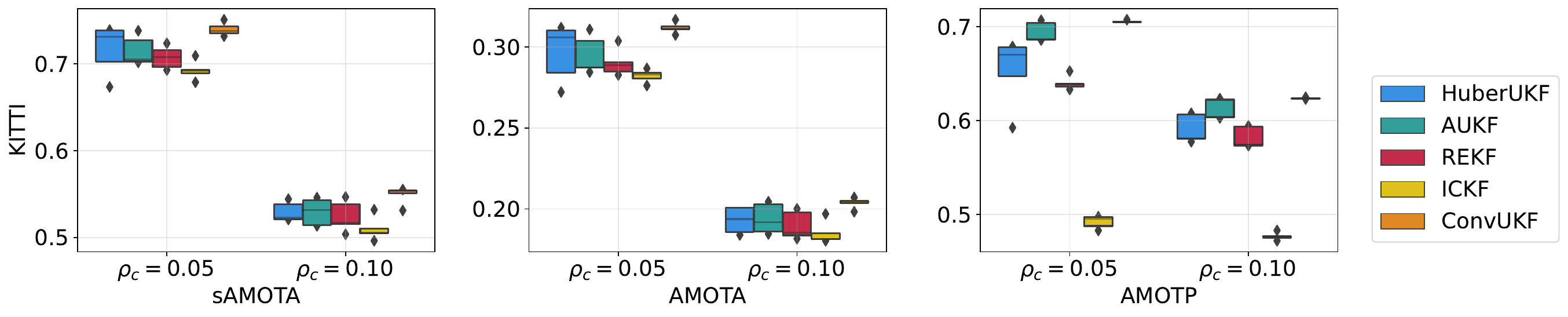}
\caption{{Boxplot illustrating sAMOTA, AMOTA, and AMOTP values obtained from HuberUKF, AUKF, REKF, ICKF, and ConvUKF with contamination probabilities $\rho_c\in\{5\%,10\%\}$ in the KITTI validation set. The circles are the data points that fall significantly outside the interquartile range.}}
\label{fig.boxenplot}
\end{figure*}

\subsection{Settings}\label{subsec.settings}
We evaluate our proposed method using the validation set of KITTI and nuScenes datasets. These datasets are widely adopted in the field of autonomous driving for their collection of raw data captured through onboard sensors during real-world driving scenarios. Specifically, we utilize raw LiDAR cloud point as our input data. To valid our method, which is employed in the filtering module, we employ the trained model PointRCNN from \cite{shi2019pointrcnn} (in KITTI dataset) and the Megvii from \cite{zhu2019class} (in nuScenes dataset) for object detection and the Hungarian algorithm \cite{kuhn1955hungarian} described in Section \ref{subsec.Preliminaries} for data association. 
For a fair comparison, all baseline algorithms utilize the same object detection and data association methodologies.

To evaluate the efficacy of our method, we conduct a comparative analysis with several established algorithms, including the UKF,  Huber unscented Kalman filter (HuberUKF) \cite{bing2018huber}, Adaptive UKF (AUKF) \cite{ge2019adaptive}, iterative extended Kalman filter (IEKF) \cite{havlik2015performance}, outlier-robust extended Kalman filter (REKF) \cite{qiu2023outlier} and improved cubature kalman filter (ICKF) \cite{qiu2020improved}, on the validation sets of the KITTI and nuScenes. 
 HuberUKF is a classical outlier-robust filtering method that incorporates the Huber loss to form a more robust cost function in the UKF. AUKF is a novelly adaptive filtering method that dynamically adjusts the covariance parameters based on estimation errors to handle complex scenes effectively. IEKF utilizes the concept of Newton's method iteration to improve nonlinear approximation effects.
 REKF and ICKF adopt factor adjustment, robust weighting, and variance calibration, demonstrating commendable robustness in practical applications. To ensure a fair benchmark comparison, the parameters for each baseline method are configured according to the specifications provided in their respective original literature. 
% More precisely, the Huber parameter for HuberUKF is set to $1$, the iterative step of IEKF denotes $10$, and the adaptive parameter $\rho = 0.9$ for AUKF. 
% Our proposed ConvUKF employs the
% adaptive update rule
% described in \eqref{eq.gamma_adaptive} for parameter $\gamma$. 

MOT challenge benchmarks employ the following metrics: scaled Average Multi-Object Tracking Accuracy (sAMOTA), Average Multi-Object Tracking Accuracy (AMOTA), Average Multi-Object Tracking Precision (AMOTP), Mostly Tracked (MT), Mostly Lost (ML), Ignored False Negatives (IFN) and Fragmentation (FRAG). For detailed definitions of these metrics, please refer to \cite{weng2019baseline, bernardin2008evaluating}.
As for hardware, our experiments were conducted on a desktop computer equipped with an Intel(R) Core(TM) i9-12900K processor and an NVIDIA 3090 Ti GPU.

% \begin{figure}[!t]
% \centering
% \includegraphics[width=1.0\columnwidth]{figure/outlier_sAMOTA.pdf}
% \caption{Plot of sAMOTA for methods with different contamination probabilities $\rho_c$ in the KITTI and nuScenes validation set. 
% }
% \label{fig.diver_pc}
% \end{figure}

\begin{table}[!t]
    \centering    \caption{\centering{\textsc{The Results on the Contaminated KITTI Validation Set.}}}
    \label{tab.val_rob_kitti}
    \begin{tabular}{lcccc}
        \toprule
        \textbf{Probability} & \textbf{Method} & \textbf{sAMOTA\%} & \textbf{AMOTA\%} & \textbf{AMOTP\%} \\
        \midrule
        \multirow{3}{*}{\textbf{$\rho_c = 0.05$}} & HuberUKF & 71.66 & 29.66 & 65.53 \\
        & AUKF & 71.69 & 29.47 & 69.07 \\
        &REKF & 70.73 & 29.00 & 63.99 \\
        &ICKF & 69.23 & 28.21 & 49.23 \\
        & ConvUKF & \textbf{73.96} & \textbf{31.17} & \textbf{70.53} \\
        \midrule
        \multirow{3}{*}{\textbf{$\rho_c = 0.10$}} & HuberUKF & 52.92 & 19.30 & 59.06 \\
        & AUKF & 52.95 & 19.40 & 61.11 \\
        &REKF & 52.41 & 18.98 & 58.18 \\
        &ICKF & 50.96 & 18.51 & 47.66 \\
        & ConvUKF & \textbf{54.99} & \textbf{20.37} & \textbf{62.37} \\

        \bottomrule
    \end{tabular}
\end{table}

\subsection{Results}
% \textbf{Evaluation Index.}
\textbf{Results on KITTI Validation Set.}
The comparative results are detailed in Table \ref{tab:performance_comparison},
which shows that ConvUKF algorithms demonstrate superior enhancements across almost all metrics compared to the baseline methods. ConvUKF achieves an impressively high sAMOTA (93.32\%) 
while maintaining the best AMOTA (45.46\%), AMOTAP (78.09\%), MT (75.68\%), and FRAG (17),
highlighting the promising potential of integrating ConvUKF into filter-based MOT methods.

Although the ML and IFN of ConvUKF are slightly inferior to those of AUKF and ICKF, ConvUKF obtains better comprehensive performance, representing a 4.26\% improvement in sAMOTA compared to the best sAMOTA achieved by REKF.

\textbf{Results on nuScenes Validation Set.}
To verify the effectiveness of ConvUKF
in different situations, we conducted the same experiment on
the nuScenes validation set, as shown in Table \ref{tab:performance_comparison_nuscenes}.
ConvUKF also demonstrates a better comprehensive performance among all the baseline methods, achieving the best sAMOTA (69.12\%), AMOTA (25.05\%), MT (42.86\%) and FRAG (1171). 
AUKF and IEKF achieve the best performance in ML and IDS, respectively, but with a relatively low sAMOTA, reflecting poor overall accuracy. This highlights that while other filtering methods may excel in specific performance indicators, they often fall short in overall tracking accuracy. In contrast, ConvUKF demonstrates a high level of overall performance, achieving strong results across all metrics.
 We notice that the improvement of  ConvUKF' sAMOTA on nuScenes (2.93\%) is lower compared with KITTI (10.27\%).
Our analysis suggests that the nuScenes dataset, with approximately 40,000 frames, contains about three times as much data as the KITTI dataset, which has 15,000 frames, and utilizes a more accurate LiDAR sensor \cite{caesar2020nuscenes}. Consequently, the proportion of outliers in nuScenes is smaller, resulting in a smaller degree of improvement with our method.

\begin{table}[!t]
    \centering
    \caption{\centering{\textsc{The Results on the Contaminated nuScenes Validation Set.}}}
    \label{tab.val_rob_nus}
    \begin{tabular}{lcccc}
        \toprule
        \textbf{Probability} & \textbf{Method} & \textbf{sAMOTA\%} & \textbf{AMOTA\%} & \textbf{AMOTP\%} \\
        \midrule
        \multirow{3}{*}{\textbf{$\rho_c = 0.05$}} & HuberUKF & 61.81 & 21.01 & 42.38 \\
        & AUKF & 55.15 & 18.20 & 39.62 \\
        & REKF & 61.18 & 20.64 & 36.33 \\
        & ICKF & 60.65 & 20.26 & 23.04 \\
        & ConvUKF & \textbf{64.37} & \textbf{22.33} & \textbf{48.84} \\
        \midrule
        \multirow{3}{*}{\textbf{$\rho_c = 0.10$}} & HuberUKF & 58.31 & 19.05 & 42.86 \\
        & AUKF & 54.93 & 17.72 & 44.46 \\
        & REKF & 56.88 & 18.85 & 42.26 \\
        & ICKF & 56.32 & 17.94 & 27.54 \\
        & ConvUKF & \textbf{59.97} & \textbf{19.92} & \textbf{47.73} \\
        \bottomrule
    \end{tabular}
\end{table}
\textbf{Validation for Robustness.}
To more rigorously validate the robustness of our method, we introduce additional data contamination as outliers to the detection results. This is achieved by masking the bounding boxes at varying probabilities $\rho_c\in\{5\%,10\%\}$ to simulate scenarios where the object detection fails to recognize targets or detection errors are significant enough to lead to incorrect association matching. 
Due to the randomness of the masking process, we repeat the experiment 5 times on the validation sets of KITTI and nuScenes.
In this experiment, we focus on the three most comprehensive tracking performance indicators: sAMOTA, AMOTA, and AMOTP.

For the KITTI validation set, we present a boxplot of sAMOTA, AMOTA, and AMOTP values in Figure \ref{fig.boxenplot} with the average metric values detailed in Table \ref{tab.val_rob_kitti}. 
HuberUKF, AUKF, and REKF exhibit similar tracking performance across different levels of outlier contamination. ICKF shows the lowest accuracy but maintains stability with very minimal variance. Notably, ConvUKF consistently demonstrates exceptional stability with a minimal variance while achieving the highest sAMOTA, AMOTA, and AMOTP.  
 In comparison, ConvUKF shows improvements of 3.17\% 5.77\% 2.11\% in sAMOTA, AMOTA, and AMOTP, respectively, at $\rho_c=0.05$ and 3.86\% 5.00\% 2.06\% at $\rho_c=0.10$.

For the nuScenes validation set, as shown in Table \ref{tab.val_rob_nus}, the performance of AUKF decreases significantly as the contamination probability increases, while HuberUKF, REKF, and ICKF which emphasize robustness, perform better.
  ConvUKF consistently outperforms other methods in terms of sAMOTA, AMOTA, and AMOTP across all levels of contamination probability, validating its robustness in handling outliers in MOT.

\begin{table}[!t]
\centering
\caption{\centering{\textsc{The Results of the Ablation Experiment on KITTI Validation Set.}}}
\label{tab:ablation experiment}
\setlength{\tabcolsep}{4pt}
\begin{tabular}{lcccc}
\toprule
\textbf{Method} & \textbf{sAMOTA\% } & \textbf{AMOTA\%} & \textbf{AMOTP\%} & \\ 
\midrule
HuberUKF & 84.42& 38.12& 61.47 \\
AUKF& 84.63& 38.02& 74.80\\ 
REKF & 89.51& 42.15& 65.23\\
ICKF & 88.34& 41.32& 45.50\\
ConvUKF ($\gamma=1\times 10^{-1}$) & 84.69& 38.31& 61.35\\
ConvUKF ($\gamma=1\times 10^{-2}$) & 91.79& 44.38& 69.60\\
ConvUKF ($\gamma=1\times 10^{-3}$) & 92.70&  45.02& 72.11\\
ConvUKF (adaptive) & \textbf{93.32}& \textbf{45.46}& \textbf{78.09}\\
\bottomrule
\end{tabular}
\end{table}

\textbf{Ablation Studies.}
We also conduct an ablation experiment on the KITTI dataset with the same settings as described in Section \ref{subsec.settings}.
The results are presented in Table \ref{tab:ablation experiment}.
Compared with the conventional ConvUKF using a carefully chosen $\gamma$, our ConvUKF with the adaptive update rule demonstrates superior performance across sAMOTA (+0.67\%), AMOTA (+0.98\%), and AMOTP (+8.28\%) metrics. 
This showcases the impact of leveraging the adaptive parameter trick on the overall performance. 

Additionally, most variants of ConvUKF outperform the baseline algorithms, further highlighting the efficacy of ConvUKF in MOT.

\textbf{Run-time Comparison.}
We compare the calculation time of filtering in MOT on the KITTI validation set (Table \ref{tab:performance_comparison}) and nuScenes validation set (Table \ref{tab:performance_comparison_nuscenes}).
Since the overall trend and size relationship are consistent across both datasets, we use the KITTI results for specific analysis.

IEKF $(0.1890 ms)$ requires the least computation time among all the methods because this method does not need to obtain sigma points, resulting in lesser computation. However, IEKF sacrifices some overall accuracy because the iterative linear approximation does not accurately approximate the nonlinear motion of objects. 
HuberUKF and ICKF, which require both sampling sigma points and Huber Loss calculation, take the longest time to calculate.
Due to the Gaussian conjugate structure similar to UKF, 
ConvUKF $(0.5726 ms)$, due to the Gaussian conjugate structure similar to UKF, shows performance comparable to UKF $(0.4225 ms)$, making it suitable for real-time object tracking applications.
% Fortunately, the disparity in computation time is minimal, typically less than $1 ms$, rendering it acceptable for real-time object tracking applications.

\begin{figure*}[!t]
\centering
\begin{subfigure}{0.49\textwidth}
\includegraphics[width=\textwidth]{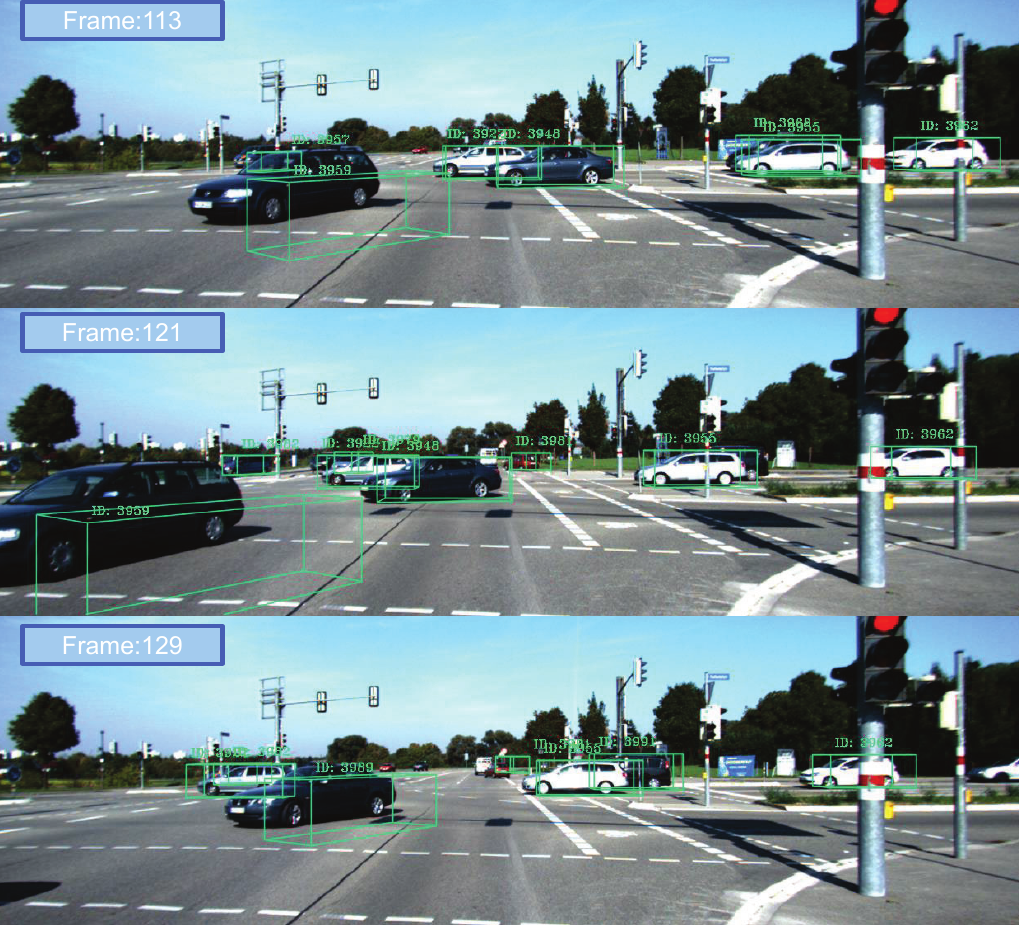}
\caption{Visualization of AUKF}
\label{fig:visual_aukf}
\end{subfigure}
\begin{subfigure}{0.49\textwidth}
\includegraphics[width=\textwidth]{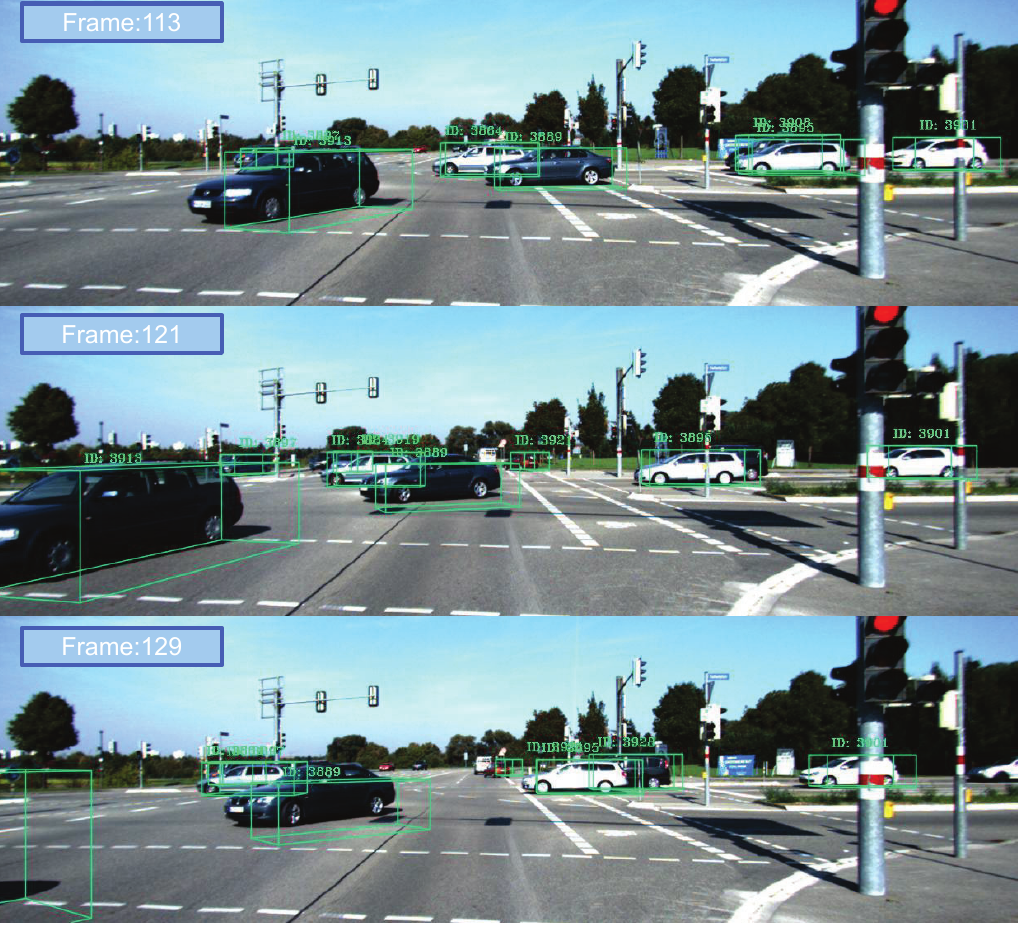}
\caption{Visualization of ConvUKF}
\label{fig:visual_conv}
\end{subfigure}
% \hfill % 
\caption{Visual comparison of MOT results between AUKF (left) and ConvUKF (right), using data from 4 continuous frames with an interval of $0.8s$ in the KITTI dataset.  For
better presentation, we visualize the results on the camera images.}
    \label{fig:pdfs}
\end{figure*}

\subsection{Visualization and Discussion}
For an intuitive analysis, we visualize the MOT results on KITTI validation set from 4 continuous frames with an
interval of $0.8s$. It is noticed that we use the raw LiDAR cloud point as the input but visualize the results on the camera images for better presentation.
As the black car in front begins to turn in Fig~\ref{fig:visual_aukf}, the AUKF's tracking (ID:3959) gradually deviates from the car's actual position until it completely loses the target. Conversely, in Fig~\ref{fig:visual_conv} ConvUKF (ID:3913) consistently maintains accurate tracking of the black car under these conditions.

In both the validation sets of KITTI and nuScenes, as well as in their data-contaminated counterparts, ConvUKF demonstrates strong performance across key metrics such as sAMOTA, AMOTP and so on. Notably, ConvUKF achieves this without significantly extending computational timeframes, ensuring practical feasibility. Moreover, ConvUKF exhibits superior robustness and minimal variance in outlier handling—a pivotal advantage for sustaining the stability and reliability of MOT systems amidst challenging conditions.

\section{Conclusion and Limitation}\label{sec.conclusion}
In this paper, we formulate a stochastic inequality to characterize the uncertainty gap between the distribution of measurements and its corresponding model for filtering. To mitigate this gap, we introduce a novel convolution-like internal process within the UKF, resulting in the ConvUKF. 
We demonstrate that ConvUKF not only maintains the conjugate structure of UKF but also obtains more robustness.
Furthermore, we prove the stability of ConvUKF in nonlinear systems with outliers, showing that ConvUKF's estimation error is bounded in the mean square sense.  Through experiments conducted on real-world datasets (KITTI and nuScenes), we showcase a performance enhancement along with acceptable computation time.

Future work will focus on how to overcome the limitations of our approach. Firstly, identify the accurate parameter $\gamma$ in ConvUKF instead of relying on the {adaptive update rule}.
We will explore employing ConvUKF in various state-of-the-art MOT algorithms \cite{li2023poly, li2024fast, wang2023camo}. 
Additionally, we plan to apply ConvUKF for MOT in real-world environments, extending beyond our current offline experiments using the dataset.

\section*{Acknowledgments}
This study is supported by National Key R\&D Program of China with 2022YFB2502901, NSF China under 52221005, Tsinghua University Initiative Scientific Research Program and Tsinghua University-Toyota Joint Research Center for AI Technology of Automated Vehicle.

{\appendices
\section{Proof of Proposition 1}
\label{appendix.Proof of Proposition 1}

From \eqref{eq.pred_x_error}, the actual prediction error covariance can be derived as
\begin{equation}
\label{eq.true_p}
\begin{aligned}
P_{t+1|t}& =\mathbb{E}[\tilde{x}_{t+1|t}\tilde{x}_{t+1|t}^\top]  \\
&=\mathbb{E}[(\alpha_tF_t(I-K_t\beta_tH_t)\tilde{x}_{t|t-1}+\xi_t)\\
&\times(\alpha_tF_t(I-K_t\beta_tH_t)\tilde{x}_{t|t-1}+\xi_t)^\top] \\
&=[\alpha_tF_t(I-K_t\beta_tH_t)]\hat{P}_{t|t-1}\\
&\times[\alpha_tF_t(I-K_t\beta_tH_t)]^\top+\Delta P^{(1)}_{t+1|t}+Q_t.
\end{aligned} 
\end{equation}
Here $\Delta P^{(1)}_{t+1|t}$ is defined as the covariance bias from expectation:
\begin{equation}
\label{eq.delta_p_1}
\begin{aligned}
\Delta P^{(1)}_{t+1|t}&=\mathbb{E}\{[\alpha_tF_t(I-K_t\beta_tH_t)\tilde{x}_{t+1|t}]\\
&\times
[\alpha_tF_t(I-K_t\beta_tH_t)\tilde{x}_{t+1|t}]^\top\}\\
&-[\alpha_tF_t(I-K_t\beta_tH_t)]\hat{P}_{t|t-1}
[\alpha_tF_t(I-K_t\beta_tH_t)].
\end{aligned} 
\end{equation}
Let $\Delta P^{(2)}_{t+1|t}$ be the difference between the real prediction error covariance $P_{t+1|t}$ and the estimated one $\hat{P}_{t+1|t}$ in \eqref{eq.prediction}:
% $\Delta P^{(2)}_{t+1|t} = P_{t+1|t} -\hat{P}_{t+1|t}$.
\begin{equation}
\label{eq.delta_p_2}
\begin{aligned}
\Delta P^{(2)}_{t+1|t} = P_{t+1|t} -\hat{P}_{t+1|t}.
\end{aligned} 
\end{equation}
Using \eqref{eq.true_p}--\eqref{eq.delta_p_2}
, we can obtain \eqref{eq.hat_P}
and \eqref{eq.Delta_P_t} with $\Delta P_{t+1|t}=\Delta P^{(1)}_{t+1|t}-\Delta P^{(2)}_{t+1|t}$.

Similarly,  we can derive the actual measurement error covariance from \eqref{eq.y_error}.
\begin{equation}
\label{eq.true_p_yy}
\begin{aligned}
P_{yy,t+1}& =\mathbb{E}[\tilde{y}_{t+1}\tilde{y}_{t+1}^\top]  \\
&=\mathbb{E}[(\beta_{t+1}H_{t+1}\tilde{x}_{t|t-1}+ \zeta_t)\\
&\times(\beta_{t+1}H_{t+1}\tilde{x}_{t|t-1}+ \zeta_t)^\top] \\
&=(\beta_{t+1}H_{t+1})\hat{P}_{t+1|t}(\beta_{t+1}H_{t+1})^\top+\Delta P^{(1)}_{yy,t+1},\\
\Delta P^{(1)}_{yy,t+1}&=\mathbb{E}[(\beta_{t+1}H_{t+1})\tilde{x}_{t+1|t}\tilde{x}_{t+1|t}^\top(\beta_{t+1}H_{t+1})^\top]\\
&-(\beta_{t+1}H_{t+1})\hat{P}_{t+1|t}(\beta_{t+1}H_{t+1})^\top.
\end{aligned} 
\end{equation}
Define $\Delta P^{(2)}_{yy,t+1}= P_{yy,t+1} -\hat{P}_{yy,t+1}$ ,  $\Delta P_{yy,t+1}= \Delta P^{(1)}_{yy,t+1} -\Delta P^{(2)}_{yy,t+1}$, and we obtain \eqref{eq.hat_Pyy} \eqref{eq.Delta_P_yy}.

% \begin{subequations}
% \begin{align}
% \hat{P}_{t+1|t}&=[\alpha_tF_t(I-K_t\beta_tH_t)]\hat{P}_{t|t-1} \nonumber \\
% &\times[\alpha_tF_t(I-K_t\beta_tH_t)]^\top+Q_t^*,
% \label{eq.hat_P}\\
% % \hat{P}_{xy,t+1}&=\hat{P}_{t+1|t}(\beta_{t+1}H_{t+1})^\top+\Delta P_{xy,t+1}+\delta P_{xy,t+1},
% % \label{eq.hat_Pxy}\\
% \hat{P}_{yy,t+1}&=(\beta_{t+1}H_{t+1})\hat{P}_{t+1|t}(\beta_{t+1}H_{t+1})^\top+\hat{R}_{t+1},
% \label{eq.hat_Pyy}\\
% % Q_t^*&=Q_t+\alpha_tF_tK_tR_t(\alpha_tF_tK_t)^\top+\Delta P_{t+1},
% % \label{eq.Q*}\\
% Q_t^*&=Q_t+\Delta P_{t+1},
% \label{eq.Q*}\\
% \hat{R}_{t+1}&=R^{'}_{t+1}+\Delta R_{t+1}+\Delta P_{yy,t+1},
% \label{eq.R*}
% \end{align}  
%  \end{subequations}

\section{Proof of Proposition 2}
\label{Appendix.Proof of Proposition 2}

Based on Assumption \ref{asp.boundeness},
we can derive the low bound of $\hat{P}_{t+1|t}$ from \eqref{eq.P_update}:
\begin{equation}\nonumber
\begin{aligned}
\hat{P}_{t+1|t} &=\hat{P}_{t+1}+K_t\hat{P}_{xy,t+1}^\top\\
&=\hat{P}_{t+1}+\hat{P}_{xy,t+1}\hat{P}_{yy,t+1}^{-1}\hat{P}_{xy,t+1}^\top\\
&\geq \hat{P}_{t+1}\geq{p}_{l}I.
\end{aligned}  
\end{equation}

Under Assumption \ref{asp.linear approximation}, we can calculate the upper bound of $\hat{P}_{t+1|t}$ as \eqref{eq.bar_p}. 
% Here $Q_t^'*$ is a similar covariance modification in \eqref{eq.Q*}, which is non-negative definite.

\begin{equation}
\label{eq.bar_p}
\begin{aligned}
\hat{P}_{t+1|t} &\leq\alpha_tF_t\hat{P}_{t+1}(\alpha_tF_t)^\top +{Q}_t+\Delta P_{t+1|t}\\
&\leq ({p}_{u}\alpha_u^{2}f_{u}^2+q_{u}+\Delta{p}_{u})I.
\end{aligned}  
\end{equation}

According to Cauchy-Schwartz inequality, we can prove
\begin{equation}
\label{eq.bar_P_xy}
\begin{aligned}
\hat{P}_{xy,t+1} &\leq \hat{P}_{t+1|t}\hat{P}_{yy,t+1}.
\end{aligned}  
\end{equation}
Substituting \eqref{eq.bar_P_xy} in \eqref{eq.K_update} gives
\begin{equation}\nonumber
\begin{aligned}
\|K_t\| &\leq \|\hat{P}_{t+1|t}\|=\sqrt{n}({p}_{u}\alpha_u^{2}f_{u}^2+q_{u}+\Delta{p}_{u})=K_u,
\end{aligned}  
\end{equation}
where $n$ is the dimension of state $x_t$.

% \begin{equation}
% \label{eq.1}
% \begin{aligned}
% \hat{P}_{t+1|t}&=\alpha_tF_t\hat{P}_{t+1}(\alpha_tF_t)^\top + Q_t^*,\\
% K_t&=\hat{P}_{t|t-1}H_t^\mathrm{T}(H_t\hat{P}_{t|t-1}H_t^\mathrm{T}+R_t^{*})^{-1}\\
% &=(I-K_tH_t)\hat{P}_{t|t-1}H_t^\mathrm{T}R_t^{*-1}=\hat{P}_tH_t^\mathrm{T}R_t^{*-1},
% \end{aligned}  
% \end{equation}

\section{Proof of Theorem 2}
\label{appendix.Proof of Theorem 2}
For analyzing the boundedness of stochastic processes, the lemma is recalled:
\begin{lemma}\label{lemma1}
(Stability of stochastic process \cite{tarn1976observers})
Assume that there is a stochastic process $\psi_t$ with transform form $V(\psi_t)$,  and real constants $\nu_{l}, {\nu}_{u}, \mu \geq 0$ and $0\leq \lambda \leq 1$ such that $\forall k$
\begin{subequations}
\begin{align}
\nu_{l}\|\psi_t\|^2\leq V(\psi_t)&\leq {\nu}_{u}\|\psi_t\|^2,\label{eq.lemma1_assumption1}\\
\mathbb{E} [ V(\psi_t)|\psi_{t-1}]-V(\psi_{t-1})&\leq\mu-\lambda V(\psi_{t-1}).\label{eq.lemma1_assumption2}
\end{align}  
\end{subequations}
Then the process $\psi_t$ is bounded in the mean square,\:i.e.,
\begin{equation}\nonumber
\begin{aligned}
\mathbb{E}\{\lVert\psi_t\rVert^2\}\leq\frac{{\nu}_{u}}{\nu_{l}}\mathbb{E}\{\lVert\psi_0\rVert^2\}(1-\lambda)^k+\frac{\mu}{\nu_{l}}\sum_{i=1}^{k-1}(1-\lambda)^i.
\end{aligned}  
\end{equation}
\end{lemma}

Define $V_{t+1}(\tilde{x}_{t+1|t})=\tilde{x}_{t+1|t}^\top \hat{P}_{t+1|t}^{-1}\tilde{x}_{t+1|t}$ and according to Prosopition \ref{prop.P_tK_t_range} , we have $\frac{\|\tilde{x}_{t+1|t}\|^2}{{p}_{u}\alpha_{u}^2f_{u}^2+q_{u}+\Delta {p}_{u}}\leq V_{t+1}(\tilde{x}_{t+1|t})^2\leq\frac{\|\tilde{x}_{t+1|t}\|^2}{{p}_{l}}$, which fulfill the first condition \eqref{eq.lemma1_assumption1} of Lemma \ref{lemma1} if we set:
\begin{equation}
\label{eq.vl_vu}
\begin{aligned}
\nu_{l}=\frac{1}{{p}_{u}\alpha_{u}^2f_{u}^2+q_{u}+\Delta {p}_{u}},\; {\nu}_{u}=\frac{1}{{p}_{l}}.  
\end{aligned}  
\end{equation}
Now let's consider :
\begin{equation}
\label{eq.EVx}
\begin{aligned}
\mathbb{E}& \{V_{t+1}(\tilde{x}_{t+1\mid t})|\tilde{x}_{t\mid t-1}\}  \\
&=\tilde{x}_{t|t-1}^\top[\alpha_tF_t(I-K_t\beta_tH_t)]^\top\hat{P}_{t+1|t}^{-1} \\
&\times[\alpha_tF_t(I-K_t\beta_tH_t)]\tilde{x}_{t|t-1} \\
&+\mathbb{E}\{(\zeta_t\alpha_tF_tK_t)^\top\hat{P}_{t+1|t}^{-1}\alpha_tF_tK_t\zeta_t|\tilde{x}_{t|t-1}\} \\
&+\mathbb{E}\{\xi_t^\top\hat{P}_{t+1|t}^{-1}\xi_t|\tilde{x}_{t|t-1}\}.
\end{aligned}  
\end{equation}

\begin{itemize}
\item 
Consider the first term of \eqref{eq.EVx}: $(\bigstar)=[\alpha_tF_t(I-K_t\beta_tH_t)]^\top\hat{P}_{t+1|t}^{-1}
[\alpha_tF_t(I-K_t\beta_tH_t)]$. 
Taking the inverse operation on it and substituting \eqref{eq.hat_P} into it, the function becomes
\begin{equation}
\label{eq.1_term}
\begin{aligned}
(\bigstar)^{-1}&=\hat{P}_{t|t-1}+[\alpha_tF_t(I-K_t\beta_tH_t)]^{-1}\hat{Q}_t\\
&\times[\alpha_tF_t(I-K_t\beta_tH_t)]\\
& \geq [1+\frac{(q_{l}+\Delta {p}_{l})I}{\hat{p}_{u}(\alpha_{u}f_{u}+\alpha_{u}f_{u}K_u\beta_{u}h_{u})^2}]\hat{P}_{t|t-1},\\
(\bigstar) &\leq [1+\frac{(q_{l}+\Delta {p}_{l})I}{\hat{p}_{u}(\alpha_{u}f_{u}+\alpha_{u}f_{u}K_u\beta_{u}h_{u})^2}]^{-1}\hat{P}_{t|t-1}^{-1}.
\end{aligned}  
\end{equation}

Conveniently,  define $\lambda$:
\begin{equation}
\label{eq.lambda}
\begin{aligned}
\lambda = 1-[1+\frac{(q_{l}+\Delta {p}_{l})I}{\hat{p}_{u}(\alpha_{u}f_{u}+\alpha_{u}f_{u}K_u\beta_{u}h_{u})^2}]^{-1},
\end{aligned}  
\end{equation}
where obviously $0\leq\lambda\leq1$.

\item Due to Assumption \ref{asp.boundeness}, the remainder of \eqref{eq.EVx} can be derived as \eqref{eq.2_term}.
\begin{equation}
\label{eq.2_term}
\begin{aligned}
\mathbb{E}&\{(\zeta_t\alpha_tF_tK_t)^\top\hat{P}_{t+1|t}^{-1}\alpha_tF_tK_t\zeta_t|\tilde{x}_{t|t-1}\} \\
&+\mathbb{E}\{\xi_t^\top\hat{P}_{t+1|t}^{-1}\xi_t|\tilde{x}_{t|t-1}\} \\
&\leq \mathbb{E}\left\{\frac{K_u^{2}\alpha_{u}^2f_{u}^2}{{p}_{l}}\mathrm{tr}\{\zeta_t^T\zeta_t\}+\frac1{{p}_{l}}\mathrm{tr}\{\xi_t^T\xi_t\}\right\} \\
&\leq\frac{K_u^{2}\alpha_{u}^2f_{u}^2
({r}_{u}+\Delta{r}_{u})}{{p}_{l}}m+\frac{q_{u}}{{p}_{l}}n=\mu.
\end{aligned}  
\end{equation}

\item Under the manipulation \eqref{eq.EVx}--\eqref{eq.2_term}, we get that $\mathbb{E}\{V_{t+1}(\tilde{x}_{t+1|t})|\tilde{x}_{t|t-1}\}-V_t(\tilde{x}_{t|t-1})\leq-\lambda V_t(\tilde{x}_{t|t-1})+\mu$, which fulfill the condition \eqref{eq.lemma1_assumption2}. Obviously, apply Lemma \ref{lemma1}, and the estimation error $\tilde{x}_{t+1|t}$ is bounded in the mean square:
\begin{equation}
\nonumber
\begin{aligned}
\mathbb{E}\{\|\tilde{x}_{t+1|t}\rVert^2\}\leq\frac{{\nu}_{u}}{\nu_{l}}\mathbb{E}\{\|x_0\rVert^2\}(1-\lambda)^t+\frac{\mu}{\nu_{l}}\sum_{i=1}^{t-1}(1-\lambda)^i.
\end{aligned}  
\end{equation}

From \eqref{eq.x_update},  we can yield $\tilde{x}_{t+1}$ is similarily bounded:
\begin{equation}
\label{eq.x_bound}
\begin{aligned}
\mathbb{E}\{\|\tilde{x}_{t+1}\|^2\}&\leq (1+f_{u}K_u\beta_{u}h_{u})\mathbb{E}\{\|\tilde{x}_{t+1|t}\|^2\}\\
&+{r}_{u}+\Delta{r}_{u}+\Delta{p}_{yy,u}.
\end{aligned}  
\end{equation}
\end{itemize}
Analyze \eqref{eq.vl_vu} \eqref{eq.lambda} \eqref{eq.2_term} and \eqref{eq.x_bound}, we can obtain how $r_u,\Delta r_u$ and $\mathbb{E}\{\|x_0\rVert^2\}$ affect the bound.
\begin{equation}
\nonumber
\begin{aligned}
\mathbb{E}\{\|\tilde{x}_{t+1}\|^2\}\leq C_1\mathbb{E}\{\|\tilde{x}_{0}\|^2\}
+C_2({r}_{u}+\Delta{r}_{u})+C_3,
\end{aligned}  
\end{equation}
where $C_1, C_2, C_3$ are constants as follows:
\begin{equation}
\label{eq.constant_bound}
\begin{aligned}
C_1&=\frac{{\nu}_{u}}{\nu_{l}}(1-\lambda)(1+f_{u}K_u\beta_{u}h_{u}),\\
C_2&=1+(1+f_{u}K_u\beta_{u}h_{u})\frac{mK_u^{2}\alpha_{u}^2f_{u}^2
}{\nu_l {p}_{l}(1-\lambda)},\\
C_3&=(1+f_{u}K_u\beta_{u}h_{u})\frac{nq_u}{\nu_l {p}_{l}(1-\lambda)}+\Delta{p}_{yy,u}.
\end{aligned}  
\end{equation}
The relationship between the bound of estimation error and $r_u$, $\Delta r_u$, and $\mathbb{E}{|x_0\rVert^2}$ is evidently linearly positive. It is notable that the analyzed bound isn't the supremacy.

\bibliographystyle{ieeetr}
\bibliography{ref}

\begin{IEEEbiography}[{\includegraphics[width=1in,height=1.25in,clip,keepaspectratio]{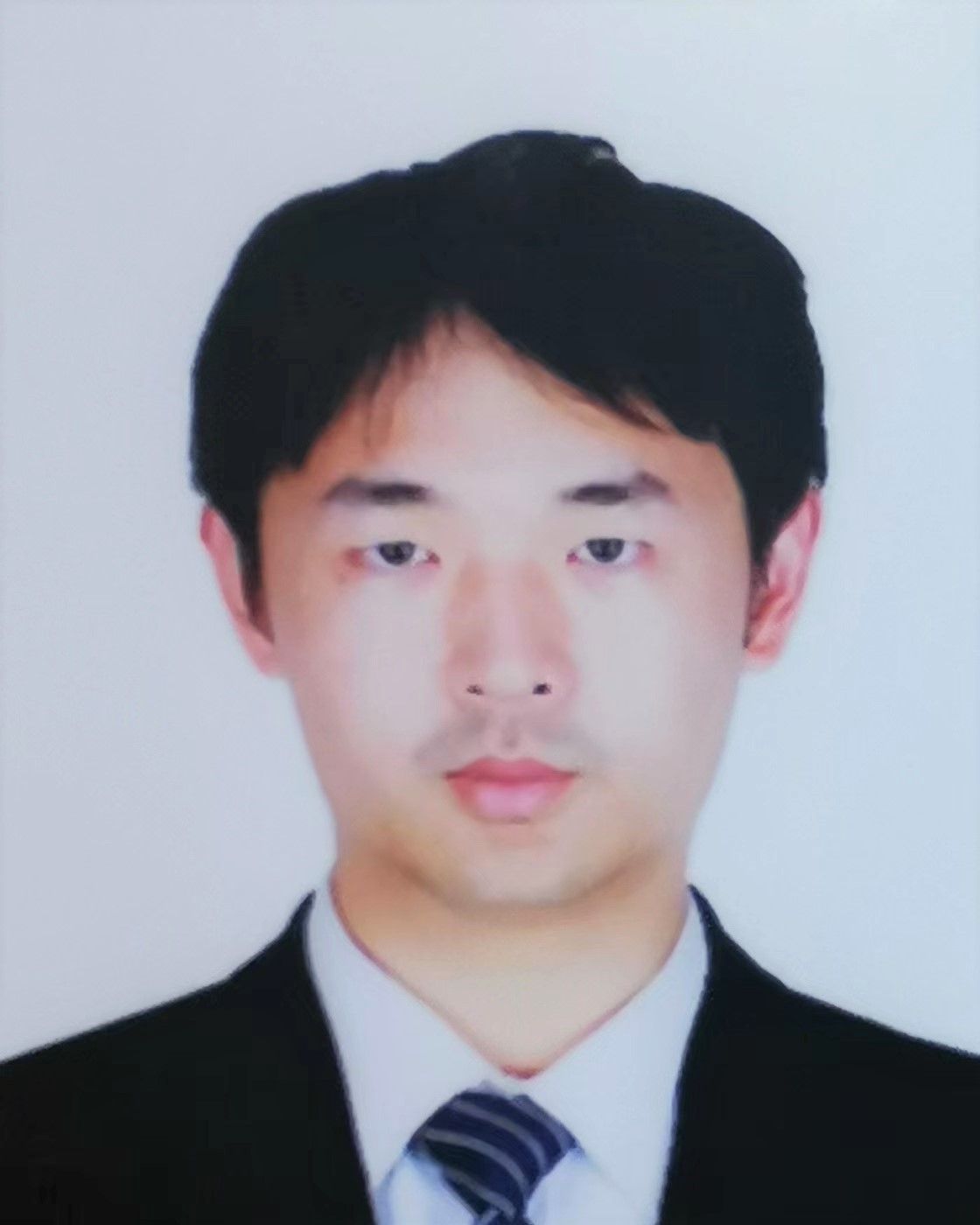}}]
{Shiqi Liu} received his B.E. degree in the School of Vehicle and Mobility from Tsinghua University, Beijing, China, in 2023.
He is currently a Ph.D. candidate in the School of Vehicle and Mobility at Tsinghua University, Beijing, China. His research interests include optimal filtering, Bayesian inference, and reinforcement learning.

\end{IEEEbiography}

\begin{IEEEbiography}[{\includegraphics[width=1in,height=1.25in,clip,keepaspectratio]{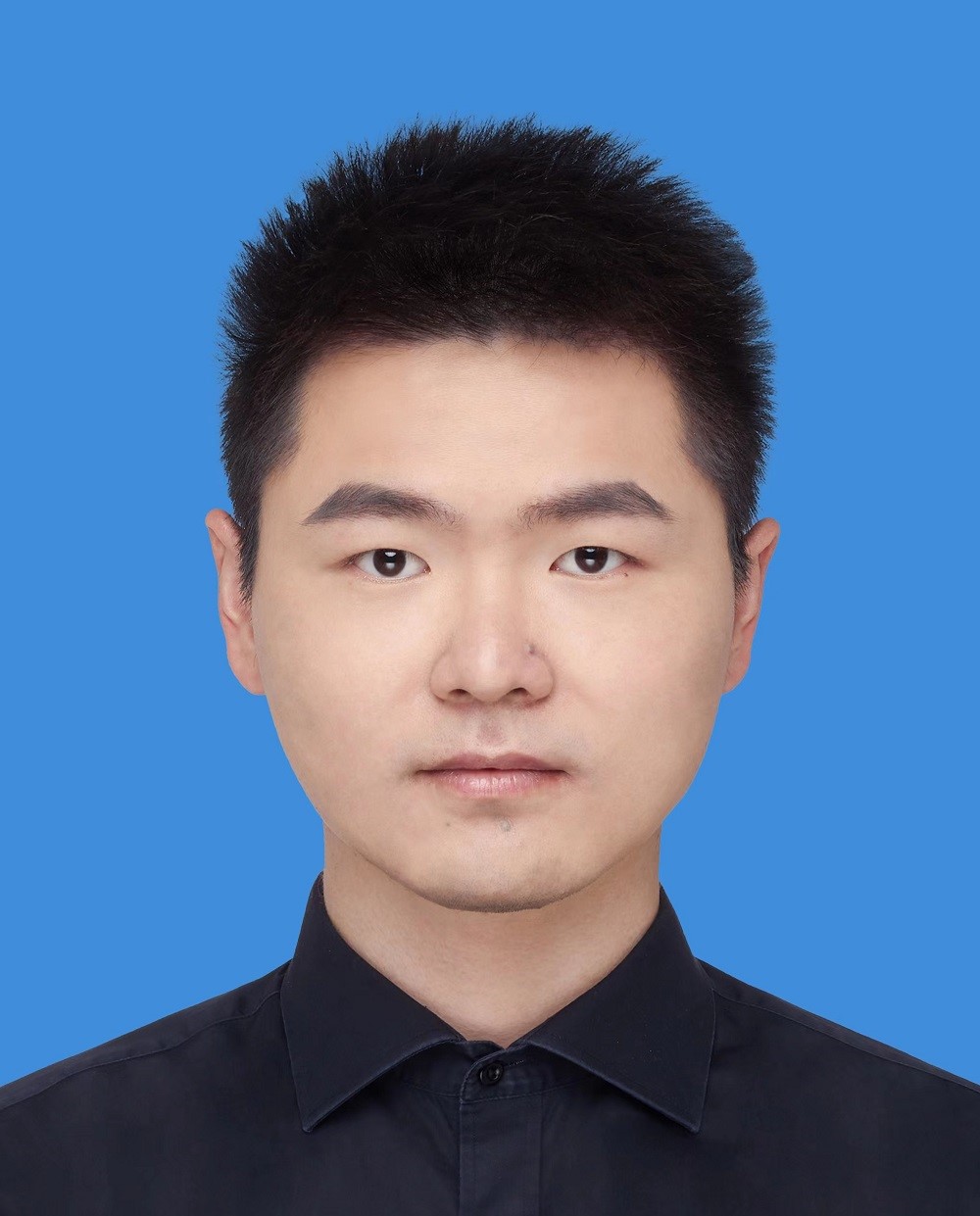}}]
{Wenhan Cao} received his B.E. degree in the School of Electrical Engineering from Beijing Jiaotong University, Beijing, China, in 2019.

He is currently a Ph.D. candidate in the School of Vehicle and Mobility, Tsinghua University, Beijing, China. His research interests include optimal filtering and reinforcement learning. He was a finalist for the Best Student Paper Award at the 2021 IFAC MECC.

\end{IEEEbiography}

\begin{IEEEbiography}[{\includegraphics[width=1in,height=1.25in,clip,keepaspectratio]{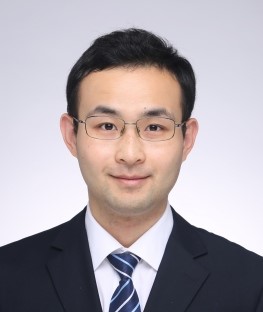}}]
{Chang Liu} (Member, IEEE) received the B.S. degrees in Electronic Information Science and Technology and in Mathematics and Applied Mathematics (double degree) from the Peking University, China, in 2011, and the M.S. degrees in Mechanical Engineering and in Computer Science, and the Ph.D. degree in Mechanical Engineering from the University of California, Berkeley, USA, in 2014, 2015, and 2017, respectively. 

He is currently an Assistant Professor with the Department of Advanced Manufacturing and Robotics, College of Engineering, Peking University. From 2017 to 2020, he was a Postdoctoral Associate with the Cornell University, USA. He has also worked for Ford Motor Company and NVIDIA Corporation on autonomous vehicles. His research interests include robot motion planning, active sensing, and multi-robot collaboration.

\end{IEEEbiography}

\begin{IEEEbiography}[{\includegraphics[width=1in,height=1.25in,clip,keepaspectratio]{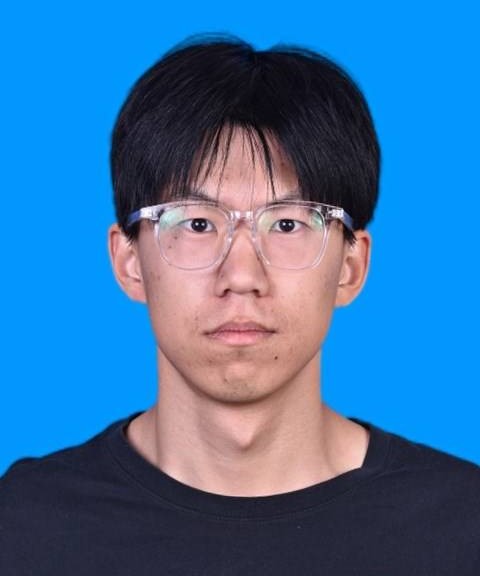}}]
{Tianyi Zhang} received his B.E. degree in Automation Science and Electrical Engineering from Beihang University, Beijing, China, in 2024. He will soon begin his Ph.D. studies in the School of Vehicle and Mobility at Tsinghua University, Beijing, China. His research interests include optimal state estimation, Bayesian inference, and reinforcement learning.
\end{IEEEbiography}

\vspace{-12cm}

\begin{IEEEbiography}[{\includegraphics[width=1in,height=1.25in,clip,keepaspectratio]{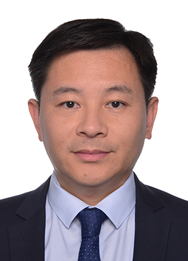}}]
{Shengbo Eben Li} (Senior Member, IEEE) received his M.S. and Ph.D. degrees from Tsinghua University in 2006 and 2009. Before joining Tsinghua University, he has worked at Stanford University, University of Michigan, and UC Berkeley. His active research interests include intelligent vehicles and driver assistance, deep reinforcement learning, optimal control and estimation, etc. He is the author of over 190 peer-reviewed journal/conference papers, and co-inventor of over 40 patents. Dr. Li has received over 20 prestigious awards, including Youth Sci. \& Tech Award of Ministry of Education (annually 10 receivers in China), Natural Science Award of Chinese Association of Automation (First level), National Award for Progress in Sci \& Tech of China, and best (student) paper awards of IET ITS, IEEE ITS, IEEE ICUS, CVCI, etc. He also serves as Board of Governor of IEEE ITS Society, Senior AE of IEEE OJ ITS, and AEs of IEEE ITSM, IEEE TITS, IEEE TIV, IEEE TNNLS, etc. 
\end{IEEEbiography}

\end{document}